\documentclass[review, anonymous]{article}

\usepackage{iclr2027_conference,times}

\iclrfinalcopy
\renewcommand{\headrulewidth}{0pt}
\makeatletter
\renewcommand{\@maketitle}{\vbox{\hsize\textwidth
{\LARGE\sc \@title\par}
\def\And{\end{tabular}\hfil\linebreak[0]\hfil
        \begin{tabular}[t]{l}\bf\rule{\z@}{24pt}\ignorespaces}%
\def\AND{\end{tabular}\hfil\linebreak[4]\hfil
        \begin{tabular}[t]{l}\bf\rule{\z@}{24pt}\ignorespaces}%
\begin{tabular}[t]{l}\bf\rule{\z@}{24pt}\@author\end{tabular}%
\vskip 0.3in minus 0.1in}}
\makeatother

\usepackage{amsmath,amssymb,amsthm}
\usepackage{graphicx}
\usepackage{booktabs}          
\usepackage{subcaption}        
\usepackage{multirow}
\usepackage{xcolor}
\usepackage{algorithm}
\usepackage{algpseudocode}
\usepackage[capitalize,noabbrev]{cleveref}   

\usepackage{wrapfig}
\usepackage{booktabs}
\usepackage{tabularx}

\graphicspath{{figures/}}

\usepackage{amsmath,amsfonts,bm}

\def\eqref#1{equation~\ref{#1}}

\def\1{\bm{1}}

\def\vmu{{\bm{\mu}}}

\def\va{{\bm{a}}}
\def\vb{{\bm{b}}}

\def\vd{{\bm{d}}}

\def\vh{{\bm{h}}}

\def\vz{{\bm{z}}}

\def\mH{{\bm{H}}}

\def\mW{{\bm{W}}}
\def\mX{{\bm{X}}}

\DeclareMathAlphabet{\mathsfit}{\encodingdefault}{\sfdefault}{m}{sl}
\SetMathAlphabet{\mathsfit}{bold}{\encodingdefault}{\sfdefault}{bx}{n}

\newcommand{\R}{\mathbb{R}}

\title{FoGConcept: Identifying Interpretable Movement Concepts for Freezing-of-Gait Detection from Multi-Scale IMU Signals}

\author{Chuntian Chi$^{1}$, John Clapham$^{2}$, GinaMari Blackwell$^{3}$, Leslie J. Cloud$^{3}$, \\
\textbf{Ingrid Pretzer-Aboff$^{3}$, Yanfu Zhang$^{1}$, Huajie Shao$^{1}$, Gang Zhou$^{1}$} \\[0.5ex]
{\normalfont $^{1}$College of William and Mary} \\
{\normalfont $^{2}$Department of Computer Science, College of William and Mary} \\
{\normalfont $^{3}$Virginia Commonwealth University}}

\begin{document}

\maketitle

\begin{abstract}
Freezing of gait (FoG) is a disabling motor symptom of Parkinson's disease, characterized by an episodic inability to initiate or sustain effective stepping. Current inertial measurement unit (IMU)-based models can detect FoG accurately, but they fail to explain their predictions in terms of human-understandable movement concepts. A major contribution of this work is to identify nine human-understandable movement concepts encoded by a multi-scale FoG model, including gait initiation, rhythmic bilateral gait, trembling in place, shuffling steps, and pivot turns. These concepts recur across subjects, are functionally linked to the model's FoG decisions, and exhibit emergent specialization across short and long temporal contexts. Repeating the concept-discovery procedure on a new FoG dataset, wider temporal contexts, and a deeper backbone recovers the same comparatively stable set of movement concepts. Accordingly, we propose FoGConcept, a mechanism-based interpretability framework for learning movement concepts from multi-scale IMU signals. Domain experts independently recognize 88.9\% of the identified concepts, while the proposed model achieves an F1 score of 94.2\% for FoG detection. Our findings show that accurate FoG detection can be understood through human-recognizable movement patterns, providing new insight into what a detection model learns to distinguish FoG from non-FoG movement.
\end{abstract}

\section{Introduction}
\label{sec:introduction}

Freezing of gait (FoG) is a disabling motor symptom of Parkinson's disease in which effective stepping is briefly interrupted despite the intention to move. FoG restricts mobility, increases fall risk, and substantially affects quality of life~\citep{bloem2004falls,lord2020freezing}. Wearable inertial measurement
units (IMUs) enable continuous gait monitoring and have supported increasingly accurate machine-learning (ML) methods for automatic FoG detection~\citep{salomon2024machine,koltermann2024gaitguard,chi2026fmfog}. However, predictive accuracy alone does not explain how these models make their decisions or what they have learned about human movement.

Existing interpretation methods for FoG detection models primarily expose low-level predictive evidence, such as an individual sensor channel, signal peak, or kinematic attributes~\citep{filtjens2021modelling,aladhaileh2025deep}. These methods show which inputs influence a prediction, but not what internal representations support the model's decision. We therefore ask: Can the internal representations supporting FoG detection be interpreted in a human-understandable manner?

To investigate this question, we introduce FoGConcept, a mechanism-based interpretability framework for learning movement concepts from multi-scale IMU signals. Specifically, we first build a multi-scale model that encodes bilateral IMU signals over short, medium, and long temporal contexts and fuses their representations for detection. A shared sparse autoencoder (SAE) then decomposes token representations across scales into sparse features. We identify human-understandable concepts from latent features through representation steering and movement-pattern characterization, and test their reproducibility through cross-subject validation. The resulting concepts are further validated by blinded domain experts.

Our analysis reveals that the model's decision-relevant features can be explained through high-level, human-understandable movement patterns. We call a recurring, human-recognizable movement pattern associated with a model-relevant feature a movement concept. We identify nine such concepts spanning localized unilateral events like single-foot strikes, locomotor transitions like the start and end of walking, sustained normal gaits, rotational coordination, PD-associated abnormal gaits like shuffling, and FoG-specific patterns such as the start of freezing and trembling in place.

\begin{wrapfigure}{r}{0.48\textwidth}
    \centering
    \vspace{-1em}
\includegraphics[width=\linewidth, trim=17cm 1cm 17cm 1cm ,clip]{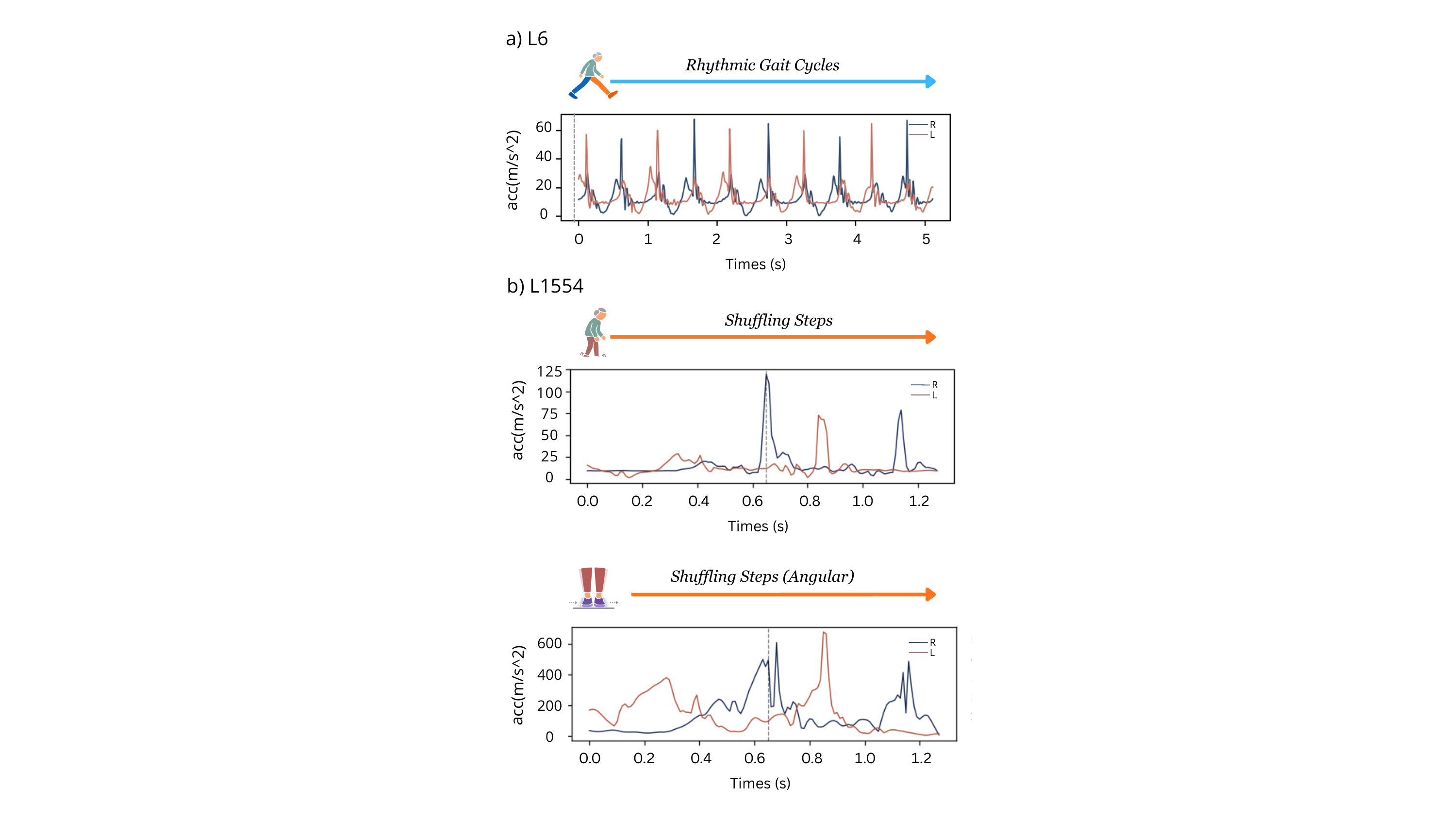}
    \vspace{-1.5em}
    \caption{
    \textbf{Examples of movement concepts.} Each panel shows a high-activation window from bilateral ankle IMUs. Red and blue denote the left and right foot, respectively. Acceleration shows the three-axis magnitude in $\mathrm{m/s^2}$; angular velocity shows the three-axis magnitude in ${}^{\circ}/\mathrm{s}$. (a) L6 captures rhythmic bilateral gait cycles with regular alternating foot activity. (b) L1554 captures shuffling steps, where repeated stepping attempts are evident in angular velocity while the corresponding acceleration is small and irregular.}
    \label{fig:intro-concepts}
    \vspace{-4em}
\end{wrapfigure}

Fig.~\ref{fig:intro-concepts} previews two examples of interpretable movement concepts. L6 represents \emph{Rhythmic Gait Cycles}, with regular acceleration peaks alternating between the two feet across successive steps. L1554 represents \emph{Shuffling Steps}, where a person moves forward with very small steps and drags their feet along the ground instead of lifting them fully~\citep{shuffle}. Here, irregular stepping attempts are evident in the gyroscope signals but produce only weak acceleration responses, consistent with limited effective progression.

Beyond their movement semantics, the discovered concepts exhibit emergent specialization across temporal contexts. Localized patterns such as the start of freezing (Fig. \ref{fig:movement-concepts}b) preferentially emerge at shorter contexts, whereas sustained or multi-cycle patterns such as rhythmic gait emerge at longer ones. The associated movement patterns also recur across held-out subjects, remain distinct from one another, and are independently recognizable by domain experts.

In summary, our contributions are:
\begin{itemize}
    \item We contribute human-understandable movement concepts underlying FoG detection, including gait initiation, single foot strikes, bilateral foot oscillation, and find emergent specialization across temporal contexts.
    \item We propose FoGConcept, the first mechanism-based interpretable framework that identifies aforementioned human-understandable concepts from multi-scale IMU data.
    \item We evaluate FoGConcept on real-world FoG data. Domain experts recognize 88.9\% of the discovered concepts, and the underlying multi-scale model achieves 94.2\% F1 for FoG detection.
\end{itemize}
\section{Related Work}
\textbf{FoG Detection and Interpretation.} Wearable IMUs are widely used for FoG detection. Earlier methods combine hand-crafted features with conventional classifiers~\citep{moore2008ambulatory}, while recent CNN-, RNN-, transformer-, and hybrid-based approaches learn temporal gait representations directly from wearable signals~\citep{delgado-teran2025ankle,koltermann2024gaitguard,chi2026fmfog}. Although these models achieve strong predictive performance, they're black boxes and remain difficult to interpret.

Existing interpretation of learning-based FoG models is predominantly input-level. Attention mechanisms highlight salient portions of sensor sequences~\citep{aladhaileh2025deep}, while layer-wise relevance propagation assigns relevance to input dimensions associated with FoG predictions
\citep{filtjens2021modelling}. These methods identify influential input evidence but do not directly characterize the movement patterns encoded in internal representations. Our work instead asks whether functionally relevant internal features correspond to human-recognizable movement patterns.

Temporal context is also important for FoG detection. Prior work either evaluates different window lengths as separate model settings or combines features computed from multiple temporal windows~\citep{kwon2014practical,Cohen_2024}. In contrast, our model jointly processes raw IMU signals at multiple input durations, learns a separate representation at each temporal scale, and fuses these representations for a single FoG prediction. This design integrates short- and long-term gait dynamics within one model, and also enables us to examine how internal movement concepts specialize across temporal contexts.

\textbf{Sparse Autoencoders for Interpretability.} Sparse autoencoders (SAEs) decompose dense neural representations into sparse latent features and have been used to identify interpretable internal structure and test feature function through interventions~\citep{cunningham2023sparse,gao2024scalingevaluatingsparseautoencoders}. SAE-based interpretability has expanded from language and vision~\citep{lim2025sparseautoencodersrevealselective}to speech~\citep{du2026sparse,koriagin2026interpretingsteeringtexttospeechlanguage} and time-series models~\citep{kalnāre2025mechanisticinterpretabilitytransformerbasedtime,oublal2026timesaecausalsparsedecoding,mishra2026dissectingchronossparseautoencoders}. These studies demonstrate that sparse feature analysis can be applied to temporal representations, but they do not examine physiological movement encoded by wearable IMUs.

The closest physiological precedent applies SAEs to EEG foundation-model representations and evaluates latent features against predefined clinical concepts, together with concept steering and spectral analysis~\citep{lehnschiøler2026mechanisticinterpretabilityeegfoundation}. In contrast, we do not begin with a predefined movement-concept vocabulary. Model-relevant latents are selected by their functional effect and subsequently evaluated for cross-subject consistency and blinded expert recognition. To our knowledge, this is the first SAE-based analysis of wearable FoG representations based on human-understandable movement concepts.

\begin{figure*}[t!]
    \centering
    \includegraphics[width=\textwidth, trim=2cm 1cm 2cm 1cm, clip]{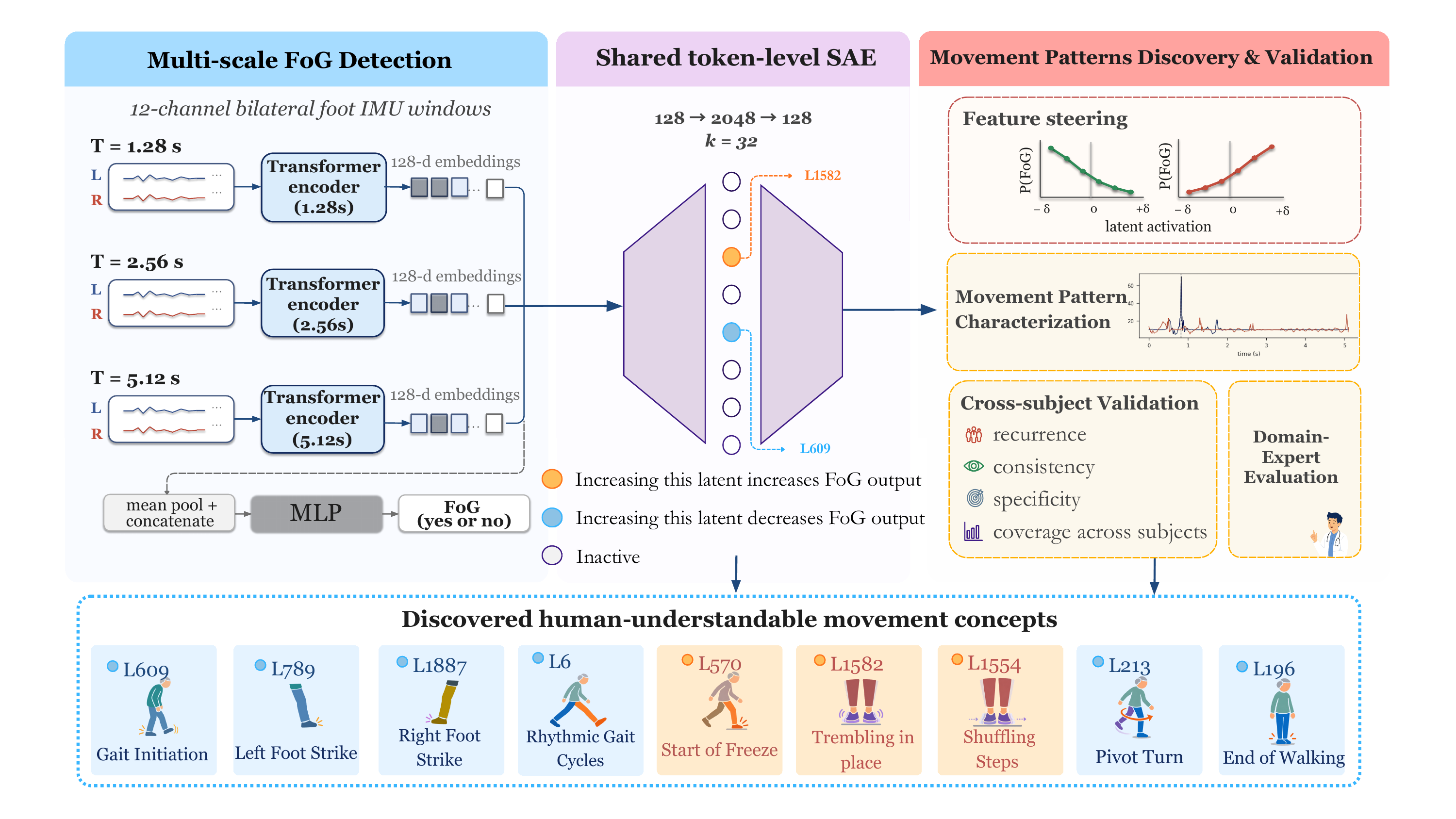}
    \vspace{-1.5em}
    \caption{
    \textbf{Overview of FoGConcept.} Bilateral ankle-IMU signals are encoded at three temporal scales and fused for FoG detection. A shared token-level SAE decomposes the encoder representations into sparse features. Model-relevant features are identified through steering, characterized from their high-activating IMU windows, and validated across subjects. The resulting movement patterns are further assessed by domain experts.
    }
    \label{fig:framework}
    \vspace{-1em}
\end{figure*}

\section{Methodology}

\subsection{Overview of FoGConcept}
\label{sec:framework}

Fig.~\ref{fig:framework} summarizes the proposed multi-scale interpretable framework, FoGConcept. A multi-scale FoG detection model encodes bilateral ankle-IMU signals at three temporal scales and combines the resulting representations for FoG detection. We then decompose its token-level representations with a shared SAE and analyze the learned SAE features through steering, movement-pattern characterization, and cross-subject validation.

The FoG model takes synchronized signals from two ankle-mounted IMUs. At each timestep, each IMU records three-axis linear acceleration and three-axis angular velocity from the left and right feet. For each prediction anchor, we extract
three aligned windows $\mX_s \in \R^{T_s\times 12}$ with $T_s\in\{128,256,512\}$ samples, corresponding to 1.28, 2.56, and 5.12s. Each window ends at the prediction anchor.

Each scale uses the Transformer configuration and masked-reconstruction pretraining procedure of FM-FoG~\citep{chi2026fmfog}, with only the input length changed across scales. The three encoders are independently parameterized. For scale $s$, the encoder produces $\mH_s =
[\vh_{s,1},\ldots,\vh_{s,T_s}]$, where each token $\vh_{s,t}\in\R^{128}$. The sequence is mean-pooled into a 128-dimensional representation. The three pooled representations are then concatenated into a 384-dimensional vector and passed to an MLP classifier to obtain $P(\mathrm{FoG})$. We use mean pooling so that token positions contribute uniformly to the pooled representation, avoiding learned position-dependent weighting in the subsequent intervention analysis. The trained FoG model is frozen for all interpretability analyses.


For interpretability, we train a shared TopK SAE~\citep{makhzani2014ksparseautoencoders,gao2024scalingevaluatingsparseautoencoders} on the token representations $\vh_{s,t}$ before pooling and cross-scale fusion. At this stage, each token still corresponds to one input timestep, allowing SAE activations to be traced back to localized regions of the IMU signal. A single SAE is trained on tokens from all three encoder streams.

For a token $\vh\in\R^d$ with $d=128$, the SAE encoder produces a sparse feature-activation vector $\vz\in\R^F$:
\begin{equation}
\vz =
\operatorname{TopK}_k\!\left(
\operatorname{ReLU}\!\left(
\mW_{\mathrm{enc}}(\vh-\vb_{\mathrm{pre}})
+\vb_{\mathrm{enc}}
\right)\right).
\label{eq:sae-encode}
\end{equation}
Here, $\mW_{\mathrm{enc}}\in\R^{F\times d}$ and $\vb_{\mathrm{enc}}\in\R^{F}$ are learned encoder parameters, and $\vb_{\mathrm{pre}}\in\R^{d}$ is a learned centering bias. The encoder centers the token representation, projects it into an $F$-dimensional feature space, and retains only the $k$ strongest positive feature responses.

The decoder maps the sparse representation back to the original token space:
\begin{equation}
\hat{\vh}
=
\mW_{\mathrm{dec}}\vz+\vb_{\mathrm{pre}}
=
\vb_{\mathrm{pre}}
+\sum_{f=1}^{F} z_f\vd_f,
\label{eq:sae-decode}
\end{equation}
where $\mW_{\mathrm{dec}}\in\R^{d\times F}$ is the learned decoder matrix and $\vd_f$ denotes its $f$th column. Thus, the reconstructed token is expressed as a weighted combination of learned decoder directions: $z_f$ specifies how strongly SAE feature $f$ is active; $\vd_f$ specifies the corresponding direction in the model's representation space. The SAE minimizes reconstruction mean squared error, with sparsity imposed directly by the TopK operation, and is frozen after training.

\subsection{Movement Pattern Discovery \& Validation}
\label{sec:pattern-discovery}

A key challenge for the proposed framework is identifying human-understandable concepts from latent features. To address this challenge, we identify movement concepts through a three-stage process. First, feature steering selects SAE features that influence the model's FoG prediction. Second, we characterize the movement pattern associated with each selected feature from its high-activation IMU windows. Third, we test whether these patterns recur and remain distinct across subjects in an independent raw-signal representation.

\paragraph{Feature Steering.} SAE decoder directions provide a direct interface for intervening on learned features and measuring their effect on downstream model behavior~\citep{du2026sparse}. We therefore use steering to identify model-relevant features before inspecting their movement semantics. For feature $f$, let $z_f(\vh_{s,t})$ denote its activation on token
$\vh_{s,t}$ and $\vd_f$ its decoder direction. We intervene only at token positions where the feature is active:
\begin{equation}
\tilde{\vh}^{(f)}_{s,t}(\alpha)=
\begin{cases}
\vh_{s,t}+\alpha\vd_f,
    & z_f(\vh_{s,t})>0,\\
\vh_{s,t},
    & z_f(\vh_{s,t})=0.
\end{cases}
\label{eq:feature-steering}
\end{equation}
The intervention therefore displaces an active token along the representation-space direction associated with feature $f$, while leaving inactive tokens unchanged. Positive $\alpha$ shifts the representation in the feature direction, negative $\alpha$ shifts it in the opposite direction, and $\alpha=0$ recovers the original representation. The edited tokens are propagated through the unchanged pooling, fusion, and classification layers.

For each feature, we evaluate a fixed set of feature-active host windows over a feature-specific range of $\alpha$. Let $\bar{p}_f(\alpha)$ denote the mean FoG probability across these windows after steering by $\alpha$. We summarize the feature's functional effect as
\begin{equation}
\Delta_f
=
\max_{\alpha\in\mathcal{A}_f}\bar{p}_f(\alpha)
-
\min_{\alpha\in\mathcal{A}_f}\bar{p}_f(\alpha),
\label{eq:functional-effect}
\end{equation}
where $\mathcal{A}_f$ is the steering grid. Thus, $\Delta_f$ measures how much the detector output changes when the representation is moved along feature $f$'s decoder direction. Features whose effect exceeds a prespecified threshold are retained as model-relevant. Feature selection is completed before movement characterization.

\paragraph{Movement Pattern Characterization.} Because SAE activations are defined at the token level, we first aggregate them to obtain one activation score per input window. For feature $f$ and window $w$ at scale $s$, we define
\begin{equation}
z_f(w)=\max_t z_f(\vh_{s,t}).
\label{eq:window-activation}
\end{equation}
We use the maximum response because the three temporal scales contain different numbers of tokens; unlike a token average, it does not dilute a localized activation in longer sequences. Windows are ranked by $z_f(w)$ to identify the examples in which feature $f$ is most strongly expressed.

Following the common SAE practice of interpreting features through maximally activating examples~\citep{bricken2023monosemanticity,lim2025sparseautoencodersrevealselective}, we inspect the highest-ranking windows. We examine the corresponding bilateral acceleration and angular velocity magnitude traces, together with their frequency content, to identify recurring movement patterns. When the high-activation examples exhibit a consistent, human-understandable movement pattern, we refer to that interpretation as the movement concept associated with feature $f$.

\paragraph{Cross-Subject Validation.}
High-activation examples alone may reflect subject-specific structure, so we test whether each characterized feature (i) recurs in recurs in subjects held out from movement characterization, (ii) preserves its
characteristic signal form, and (iii) remains distinct from other features. Each window is represented by four magnitude trajectories: left- and right-foot acceleration and angular velocity. Each trajectory is resampled to 32 points, and the four trajectories are concatenated into a fixed 128-dimensional descriptor $\bm{\phi}(w)\in\R^{128}$. For feature $f$, windows are labeled active ($z_f(w)>0$) or inactive, with inactive controls matched by temporal scale; all probes and prototypes are constructed on characterization subjects and applied without refitting to evaluation subjects.

\emph{Recurrence} trains a logistic probe $\bm\phi(w)\!\rightarrow\!\{\text{active},\text{inactive}\}$ and computes held-out ROC-AUC from the active/inactive labels and predicted active-class scores, testing whether feature-associated signal structure reappears across subjects. \emph{Prototype consistency} uses no classifier: active characterization windows are averaged into a fixed prototype, each evaluation window is scored by cosine similarity to it, and ROC-AUC tests whether active windows are more prototype-like than inactive controls.
\emph{Specificity} trains pairwise probes $\bm\phi(w)\!\rightarrow\!\{f_i,f_j\}$ and evaluates held-out ROC-AUC, testing whether different feature-associated patterns remain distinguishable. Overlapping windows are assigned to the feature with stronger activation. Full definitions and implementation details are provided in Appendix~\ref{app:crosssubject}.
\section{Evaluation}
\subsection{Experimental Settings}
\label{sec:experimental-settings}


\paragraph{Dataset and FoG detection setup.}
We use the VCU FoG-IMU dataset~\citep{chi2026fmfog}, containing bilateral ankle-IMU recordings from 23 participants with Parkinson's disease (237.9 min; 528 annotated FoG episodes) sampled at 100Hz (128 samples for a 1.28s window). FoG detection is evaluated using 23-fold leave-one-subject-out (LOSO) cross-validation. For each prediction anchor, the 2.56s window defines the label: a sample is positive if FoG occurs within the final 30\% of the window and negative otherwise~\citep{koltermann2024gaitguard,chi2026fmfog}. The aligned 1.28, 2.56, and 5.12s windows end at the same prediction anchor and are fused to predict this common label. We re-run all external baselines with 2.56s inputs using the same preprocessing, labeling rule, and LOSO folds, and additionally evaluate a single-scale 2.56s version of our model. Implementation details are provided in Appendix~\ref{app:implementation}.

\paragraph{Interpretability setup.}
For interpretability, the multi-scale model and shared SAE are trained on all 23 participants and frozen before analysis. We use $d_{\mathrm{SAE}}=2048$ and $k=32$; sensitivity to SAE configuration is reported in Appendix~\ref{app:sae}. The 10/13 subject split is used only for post-hoc concept characterization and validation: 10 participants form the characterization group and 13 disjoint participants form the evaluation group. For steering, host windows must activate the target feature and have $P(\mathrm{FoG})\in[0.15,0.85]$. For feature $f$, we sweep $\alpha\in[-3m_f,3m_f]$, where $m_f$ is its median nonzero activation, and retain features with $\Delta_f\geq0.10$.

\paragraph{Metrics.}
FoG detection is evaluated using F1, accuracy, precision, and recall, with F1 as the primary metric. Cross-subject recurrence, prototype consistency, and pairwise specificity are evaluated using ROC-AUC on the disjoint evaluation subjects; subject coverage is reported descriptively. SAE representation fidelity and configuration sensitivity are reported in Appendix~\ref{app:sae}. Because prior interpretable FoG methods provide input-level attributions rather than discovered internal concepts, there is no directly comparable concept-discovery baseline; we therefore evaluate interpretability through functional, cross-subject, and expert validation.

\subsection{Human-Understandable Movement Concepts}
\label{sec:movement-concepts}

\begin{figure*}[t]
    \centering
    \includegraphics[width=\textwidth, trim=1cm 7cm 23cm 7cm, clip]{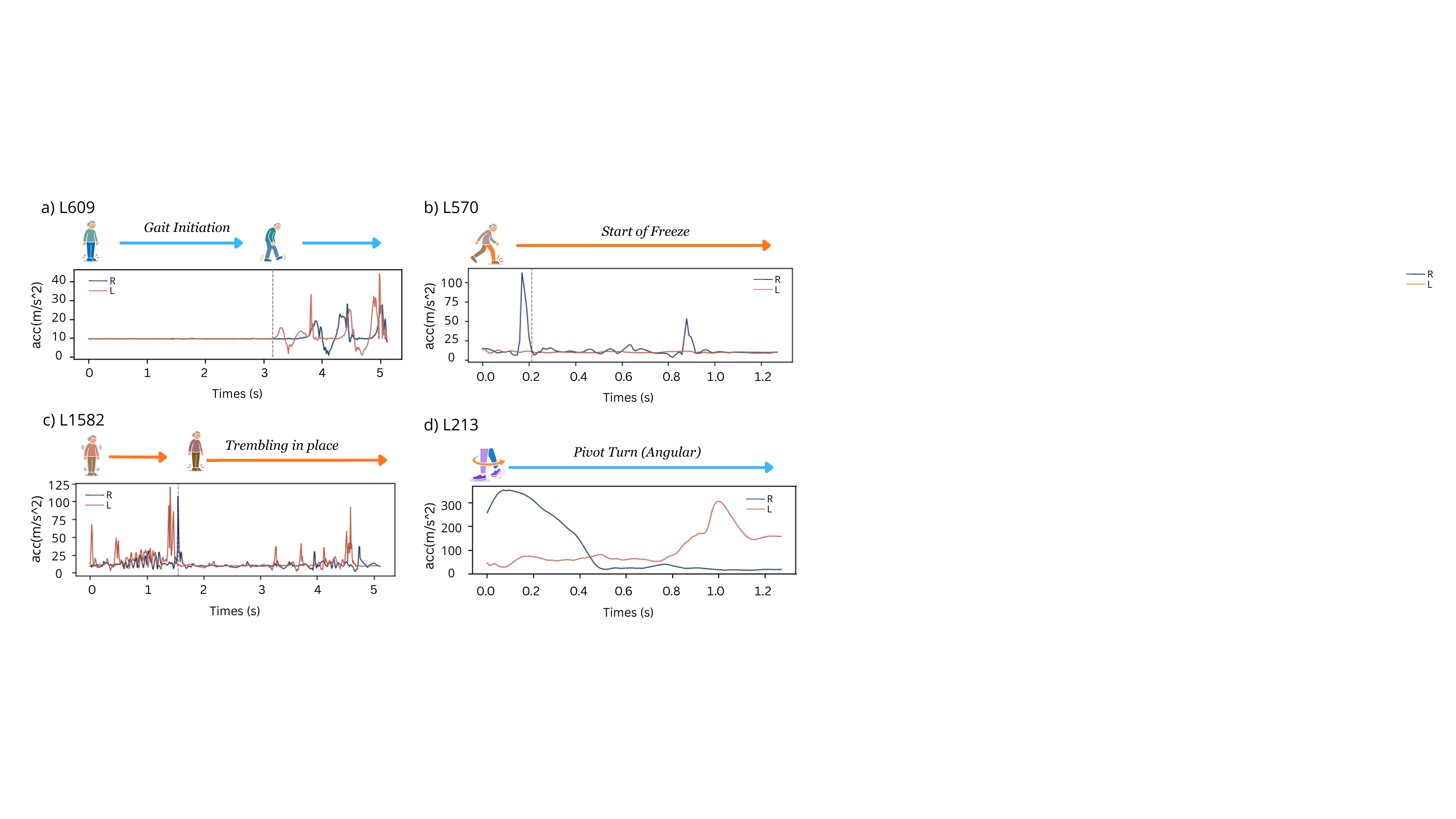}
    \vspace{-1.5em}
    \caption{
    \textbf{Representative human-understandable movement concepts.} Red and blue denote the left and right foot, respectively. Acceleration magnitude is shown for (a) L609 \emph{Gait Initiation}, from standing still to walking, (b) L570 \emph{Start of Freeze}, the transition from a step into FoG and (c) L1582 \emph{Trembling in Place}, represented by rapid lower-limb movement without stable gait cycles; angular-velocity magnitude is shown for (d) L213 \emph{Pivot Turn}, the rotational movement of the two feet when turning around. Vertical dashed lines indicate the temporal location of maximum feature activation.}
    \label{fig:movement-concepts}
    \vspace{-1em}
\end{figure*}

We first examine the movement patterns represented by the model relevant SAE features. Fig.~\ref{fig:intro-concepts} and Fig.~\ref{fig:movement-concepts} collectively show six representative concepts identified from their high-activation windows (complete atlas in Appendix \ref{app:atlas}). Each trace shows the Euclidean magnitude across the three sensor axes. Acceleration is reported in $\mathrm{m/s^2}$ and angular velocity in ${}^{\circ}/\mathrm{s}$; red and blue denote the left and right foot, respectively. Because acceleration magnitude includes gravity, a stationary foot appears as a nearly constant trace around $9.8\,\mathrm{m/s^2}$. We show the modality that most clearly exposes each pattern, and both modalities when their relationship is informative. Vertical dashed markers indicate the temporal location of maximum feature activation where shown.

These concepts do not all represent FoG itself; rather, they span ordinary gait, PD-associated gait abnormalities, transitions into FoG, and recognized FoG phenomena that contribute to the model's FoG decisions. First, they capture distinct structure within ordinary gait. L609 (\emph{Gait Initiation}; Fig.~\ref{fig:movement-concepts}a) shows both feet remaining close to a stationary acceleration baseline before the feature activates. At the activation point, the traces depart from this baseline and bilateral foot movement begins, followed by repeated stepping activity. The feature therefore marks the transition from standing to walking. L6 (\emph{Rhythmic Gait Cycles}; Fig.~\ref{fig:intro-concepts}a) captures the sustained normal walking pattern. Across the window, acceleration peaks recur at a regular cadence and alternate between the left and right feet, producing the bilateral cycles characteristic of continuous walking. L609 and L6 separate the onset of locomotion from the rhythmic organization of established gait.

L570 (\emph{Start of Freeze}; Fig.~\ref{fig:movement-concepts}b) captures the transition from an effective step into FoG. The window contains a single prominent right-foot acceleration spike, corresponding to the final effective step. Maximum feature activation occurs on the falling edge of this spike, precisely as the step terminates. For most of the following interval, both feet return close to near-stationary, indicating little foot movement corresponding with the annotated FoG period. L570 therefore localizes the boundary at which effective stepping ends and freezing begins. This transition emerges without explicit supervision for gait phase or FoG-onset timing.

L1582 (\emph{Trembling in Place}; Fig.~\ref{fig:movement-concepts}c) captures a different organization of movement around FoG. The acceleration traces contain dense, rapid fluctuations from both feet, punctuated by larger transient peaks. Unlike L6, however, these fluctuations do not recur at a stable cadence or form a consistent alternating left-right sequence. A near-stationary interval is also visible around the middle of the example during freezing, with irregular movement appearing around it. Thus, the lower limbs can remain highly active without forming the coordinated gait cycles required for effective walking. This pattern is consistent with trembling-in-place FoG, characterized by rapid leg movements with little or no effective forward progression \citep{nutt2011freezing}. The contrast between L6 and L1582 shows that the representation distinguishes organized rhythmic gait from high-frequency movement that does not resolve into effective stepping.

L1554 (\emph{Shuffling Steps}; Fig.~\ref{fig:intro-concepts}b) represents another form of disrupted gait and requires the combined analysis of both acceleration and angular velocity. In PD, a shuffling gait is characterized by very short, low-clearance steps that produce limited forward progression while stepping continues \citep{nutt2011freezing}. This behavior is reflected in the two IMU modalities. The acceleration traces remain relatively close to stationary and lack the regular, alternating peaks seen during normal walking, indicating weak and irregular translational foot dynamics. In contrast, the gyroscope shows attempted angular motion that alternates between the two feet, demonstrating that the feet continue to rotate and try to participate in a stepping sequence. In several intervals, substantial rotation (angular excursion) of one foot  is accompanied by only a small acceleration response. These signals indicate attempts in stepping with limited effective progression, consistent with the short, poorly advancing steps of shuffling.

L1554 and L1582 therefore represent clinically distinct movement patterns: shuffling is a PD-associated gait abnormality, whereas trembling in place is a recognized FoG phenotype. Shuffling preserves bilateral walking attempts, most clearly visible in angular velocity, but the associated acceleration is weak and irregular, whereas trembling in place contains rapid lower-limb activity without stable gait cycles. The representation thus separates distinct organizations of impaired movement, and does not collapse different abnormal PD gait manifestations into a generic abnormal-motion
feature.

Finally, L213 (\emph{Pivot Turn}; Fig.~\ref{fig:movement-concepts}d) captures bilateral coordination in rotational motion. The right-foot gyroscope first forms a large, smooth, bell-like excursion in angular-velocity while the left foot remains stable. As the right-foot excursion returns toward baseline, the roles reverse: the left foot develops a similarly broad rotational excursion while right-foot angular motion remains low. These two sequential rotational bouts are consistent with one foot rotating while the contralateral foot acts as support, followed by an exchange of roles as the turn progresses. 


Beyond their movement semantics, the recovered features exhibit emergent temporal-scale organization. Although the shared SAE receives no explicit scale identifier, five of the nine concepts have their highest scale-normalized firing rate in the 1.28s stream, while four are dominated by the 5.12s
stream (Appendix~\ref{app:function}). The short-context concepts include Left and Right Foot Strike, Start of Freeze, Shuffling Steps, and Pivot Turn, whereas Gait Initiation, Rhythmic Gait Cycles, Trembling in Place, and End of Walking prefer the long context. This organization aligns with movement duration: localized events and transitions preferentially occur at shorter contexts, whereas sustained or multi-cycle patterns occur at longer contexts.

\subsection{Evidence for Functional Relevance and Interpretability}
\label{sec:interpret-evidence}

We next test whether the recovered movement concepts are functionally involved in the model's FoG decisions and validate that their associated signal patterns are not subject-specific.

\begin{figure*}[t!]
    \centering
    \includegraphics[width=\textwidth, trim=0cm 4cm 0cm 0cm, clip]
    {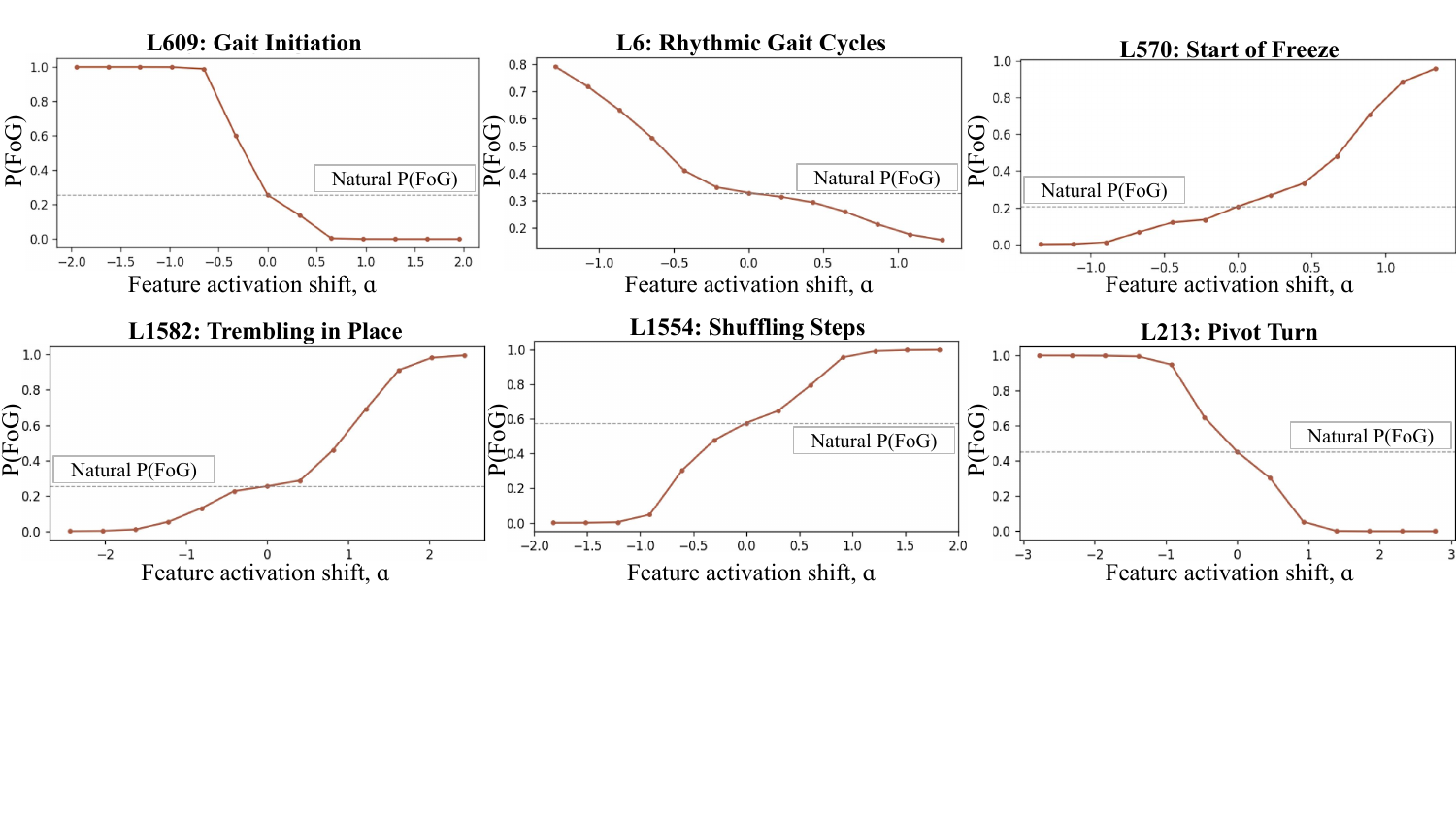}
\vspace{-1em}    \caption{\textbf{Representative feature-steering responses for movement concepts.} Each curve shows the mean predicted $P(\mathrm{FoG})$ as the activation of one SAE feature is shifted by $\alpha$ along its decoder direction. The dashed line marks the prediction at the feature's natural activation, with $\alpha=0$ corresponding to the unmodified representation. Positive steering of L570, L1582, and L1554 increases the FoG prediction, whereas positive steering of L609, L6, and L213 decreases it.}
    \label{fig:steering}
    \vspace{-1.5em}
\end{figure*}

\paragraph{Functional relevance.} Fig.~\ref{fig:steering} shows steering effects for six representative concepts (complete steering effects in Appendix \ref{app:function}). Positive steering of L570, L1582, and L1554 increases $P(\mathrm{FoG})$, whereas steering L609, L6, and L213 decreases it. Several interventions move the prediction across a large fraction of its $[0,1]$ range, showing that individual sparse features can have substantial effects on the frozen classifier. Steering directions show features for both FoG episodes and normal gaits. 

\paragraph{Random-intervention control.} We repeat steering with 1,000 norm-matched random directions while preserving the same host windows, token positions, and intervention magnitudes, and steering along the concepts' directions produce significantly larger effects than the random baseline (full details in Appendix~\ref{app:random-steering}), indicating that the steering effects are direction-specific, not a generic consequence of perturbing the representation.

\paragraph{Cross-subject generality and distinctness.}
The discovered movement patterns generalize beyond the subjects used for their characterization. Recurrence AUC ranges from 0.90 to 0.99 (median 0.96). Thus, for the median concept, a feature-active window receives a higher recurrence score than a matched inactive control in approximately $96\%$ of positive--negative comparisons, showing that feature-associated raw-signal structure remains strongly distinguishable from inactive controls. Prototype-consistency AUC ranges from $0.88$ to $1.00$ (median $0.95$), showing that held-out active windows also remain more similar than controls to the fixed pattern estimated from characterization subjects. The concepts occur in 15--23 participants (median 19), not confined to isolated individuals.

The concepts also remain distinct from one another: pairwise specificity has a median held-out AUC of 0.88, including 0.92 among same-scale pairs and 0.86 across scales. These results show that the recovered concepts correspond to movement patterns that recur across individuals, preserve their characteristic form, and remain distinguishable from other discovered
patterns. Full per-concept and pairwise results are provided in
Appendix~\ref{app:crosssubject}.

\subsection{FoG Detection Performance}
\label{sec:fog-performance}

\begin{wraptable}{r}{0.48\textwidth}
\centering
\vspace{-3em}
\caption{\textbf{FoG detection LOSO performance.} All baselines are re-run with 2.56s inputs under the common preprocessing, labeling, and LOSO protocol. The best external result is underlined and the best overall result is bold.}
\label{tab:fog-performance}

\small
\setlength{\tabcolsep}{3pt}
\renewcommand{\arraystretch}{0.9}

\begin{tabularx}{\linewidth}{@{}>{\raggedright\arraybackslash}X r@{}}
\toprule
Method & $F_1$ (\%) $\uparrow$ \\
\midrule
Dual-Level~\citep{duallevel} & 65.1 \\
Huang et al.~\citep{HUANG2024105613} & 80.4 \\
Borz\`i et al.~\citep{BORZI2025110530} & 81.8 \\
Delgado-Ter\'an et al.~\citep{delgado-teran2025ankle} & 84.0 \\
Xia et al.~\citep{XIA2024105765} & 84.5 \\
MCT-Net~\citep{mctnet} & 86.6 \\
Abbasi et al.~\citep{abbasi2025deep} & 87.2 \\
Al-Adhaileh et al.~\citep{aladhaileh2025deep} & 87.3 \\
Gait-Guard~\citep{koltermann2024gaitguard} & 88.2 \\
Contest (best)~\citep{salomon2024machine} & \underline{88.7} \\
\midrule
Ours, single-scale (2.56\,s) & 92.0 \\
\textbf{Ours, multi-scale} & \textbf{94.2} \\
\bottomrule
\end{tabularx}
\vspace{-2em}
\end{wraptable}

Table~\ref{tab:fog-performance} shows that the multi-scale model achieves the best FoG detection performance, with an F1 of 94.2\%. This exceeds the strongest baseline, the best-performing contest submission at 88.7\%, by 5.5\%. The model also achieves 94.3\% accuracy, precision, and recall, indicating balanced performance across metrics.

The single-scale model achieves 92.0\% F1, outperforming every external baseline. Jointly modeling the aligned 1.28, 2.56, and 5.12s contexts further improves F1 by 2.2 points. These results show that the multi-scale formulation provides predictive information beyond the 2.56s representation alone.

The movement concepts analyzed in this work therefore arise from the internal representations of a high-performing FoG model. The multi-scale gain further establishes that information across temporal contexts is useful for the detection task, complementing our analysis of how model-relevant movement features are organized across temporal scales.

\subsection{Reproducibility Across Datasets and Model Settings}
\label{sec:reproducibility}

We test whether FoGConcept can be generalized to identify movement concepts across datasets and model settings. We repeat the complete pipeline on FoG-STAR~\citep{fogstar}, an independent dataset of 22 PD participants collected with different IMU hardware, sites, tasks, and source sampling rate. Using bilateral ankle sensors, all nine movement-concepts from the primary VCU analysis recur (see Appendix~\ref{app:robustness} for full concept examples and SAE diagnostics.

\begin{figure}[t]
    \centering
    \includegraphics[width=\textwidth, trim=1cm 8cm 1.5cm 8.5cm, clip]{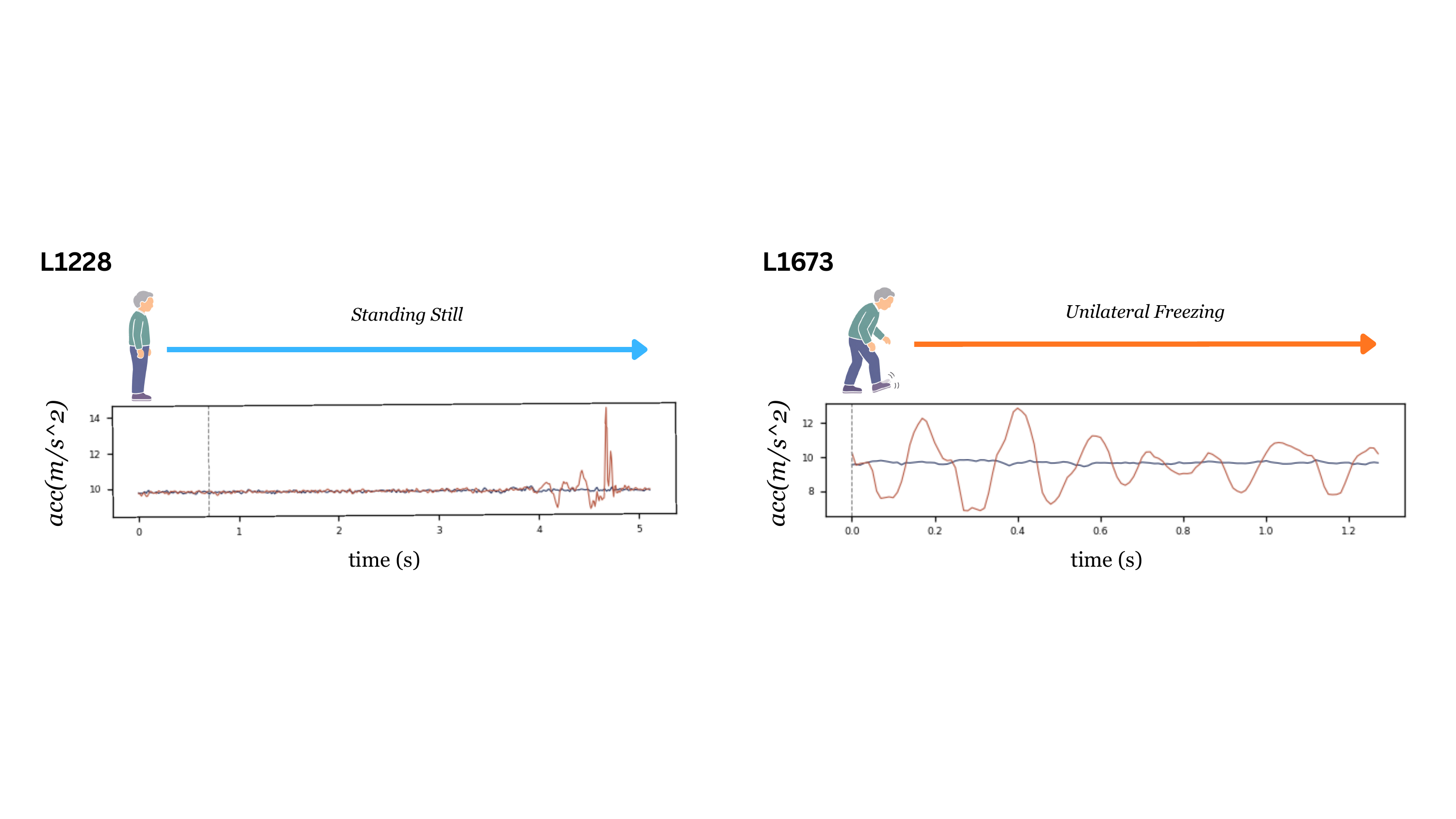}
    \vspace{-1.5em}
    \caption{\textbf{Additional movement concepts exposed by deeper backbones.} (a) Unilateral Freezing from the 12-layer VCU model, one foot exhibits repeated movement attempts while the other foot remains unable to move. (b) Standing Still from the 12-layer FoG-STAR model, both feet remain stationary in non-freezing episodes, indicating voluntary stops during walking. Red and navy denote the left and right foot; dashed lines indicate maximal feature activation.}
    \label{fig:additional-concepts}
\end{figure}

We further test on wider timescales with 10.24s windows and independently increase the Transformer backbone from 6 to 12 layers. Both wider-context models recover all nine original concepts. Both 12-layer models recover seven of the nine concepts, while also exposing additional interpretable patterns. On VCU, the deeper model identifies \emph{Unilateral Freezing}, in which one foot repeatedly attempts to move while the other foot remains planted or nearly motionless. On FoG-STAR, the deeper model identifies \emph{Standing Still}, which represent voluntary, non-freezing stops during walking. Representative examples of these additional concepts are shown in Fig.~\ref{fig:additional-concepts}.

\begin{table}[t]
\centering
\small
\caption{\textbf{Reproducibility across datasets and model settings.}
Recovered denotes recurrence of a movement-concept family from the primary VCU analysis.}
\label{tab:concept-reproducibility}
\setlength{\tabcolsep}{4pt}
\begin{tabular}{p{0.29\linewidth}p{0.30\linewidth}cc}
\toprule
Setting & Main change & Recovered & Additional concept \\
\midrule
FoG-STAR, 6-layer & Dataset / hardware & 9/9 & -- \\
FoG-STAR, wider & Temporal contexts & 9/9 & -- \\
VCU, wider & Temporal contexts & 9/9 & -- \\
FoG-STAR, 12-layer & Backbone depth & 7/9 & Standing Still\\
VCU, 12-layer & Backbone depth & 7/9 & Unilateral Freezing \\
\bottomrule
\end{tabular}
\end{table}

Across these settings, the exact concept set varies slightly, but most movement-concepts remain stable. These results support FoGConcept as a reproducible interpretability pipeline that recovers a comparatively stable set of human-understandable movement concepts, while allowing representation-dependent variation in the concepts exposed by different model configurations.

\subsection{Domain-Expert Evaluation}
\label{sec:expert-evaluation}

We further assess whether the movement concepts are independently recognizable to domain experts. Three experts with a combined 20+ years of experience in Parkinson's disease, FoG, and gait analysis independently reviewed the activation examples associated with the nine concepts and described the recurring movement, without access to one another's responses. A concept is considered recognized when at least two of the three independent descriptions identify the same underlying movement.

Experts recognized 8 of 9 concepts (88.9\%), including five with unanimous agreement. To test whether this level of agreement could arise from chance label alignment, we independently permuted the expert concept assignments and recomputed majority agreement. Experts agreement was significantly greater than the random-label distribution (permutation test, $p<0.001$). The free-text responses also recovered characteristic movements. For example, L609 was described as ``standing up and walking,'' corresponding to Gait Initiation, while L213 was described as ``turning in a circle,'' same as Pivot Turn.

After the independent review, differences in terminology or movement specificity were resolved by reviewing the corresponding source videos and FoG annotations, and with discussion, producing the final consensus labels reported in Fig.~\ref{fig:framework}.
\section{Conclusion and Limitations}
\label{sec:conclusion}
We identify human-understandable movement concepts underlying a multi-scale FoG model, spanning localized events and sustained gait patterns. Steering, cross-subject validation, expert evaluation and experiments across datasets and model configurations show that these concepts are functionally linked to FoG predictions, recur across individuals, and are human-recognizable. However, some limitations remain to be addressed. As training SAEs on large-scale data requires non-trivial computational resources, our analysis is limited to a small expert cohort. The discovered concepts are not necessarily exhaustive, as we focus on representative, clearly observable features, not systematically characterizing the full latent space. Finally, steering operates only in representation space since there is no reliable generative decoder for IMUs. Future work should test concept generality across models and datasets, and develop signal-space counterfactuals.

\section*{Ethics Statement}
The wearable data used in this study were collected under an Institutional Review Board (IRB)-approved human-subject research protocol with informed consent and were analyzed in de-identified form. Our analyses are intended to improve understanding of learned representations in freezing-of-gait models; the discovered movement concepts should not be interpreted as independently validated clinical biomarkers or diagnostic categories.

\section*{Reproducibility Statement}
Sections 3.1–3.2 specify the multi-scale model, SAE, feature-steering, movement-characterization, and cross-subject validation procedures; Sec. 4.1 describes the dataset splits, labeling protocol, steering settings, and evaluation metrics. Appendices A–E provide implementation details, the complete concept atlas and steering analysis, cross-subject validation definitions and results, and SAE reconstruction fidelity.

\section*{AI Use Statement}
Generative AI tools were used to assist with language editing and presentation refinement. All experiments and analyses were conducted by the authors. All AI-assisted content was reviewed and verified by the authors.


\bibliographystyle{iclr2027_conference}
\bibliography{main}

%
%

\clearpage
\appendix

\section{Model Implementation Details}
\label{app:implementation}

\begin{table}[h!]
\centering
\small
\caption{\textbf{Model Implementation details.}}
\label{tab:app-implementation}
\setlength{\tabcolsep}{5pt}
\begin{tabular}{ll}
\toprule
Component & Setting \\
\midrule
\multicolumn{2}{l}{\textit{Encoder and pretraining}} \\
Input channels & 12 bilateral IMU channels \\
Sampling rate & 100\,Hz \\
Temporal lengths & 128 / 256 / 512 samples \\
Input projection & $12\rightarrow128$ \\
Hidden dimension & 128 \\
Transformer blocks & 6 \\
Attention heads & 8 (head dimension 16) \\
Feed-forward network & $128\rightarrow512\rightarrow128$ \\
FFN activation & ReLU \\
Block structure & Residual connections + LayerNorm \\
Masked fraction & 30\% \\
Reconstruction head & $128\rightarrow12$ \\
Pretraining loss & MSE on masked positions \\
Pooling & Mean pooling \\
Cross-scale fusion & Concatenate pooled representations \\
Classifier & $384\rightarrow128\rightarrow2$ \\
Classifier dropout & 0.4 \\
Optimizer & AdamW with cosine annealing \\
\midrule
\multicolumn{2}{l}{\textit{Sparse autoencoder}} \\
SAE dimensions & $128\rightarrow2048\rightarrow128$ \\
TopK sparsity & $k=32$ \\
Training objective & Reconstruction MSE \\
Additional $\ell_1$ penalty & None \\
Optimizer & AdamW \\
Learning rate & $3\times10^{-4}$ \\
Validation split & 5\% \\
Decoder normalization & Unit-norm during training \\
Inactive-feature handling & Periodic resampling \\
Alive-feature threshold & Firing rate $>10^{-3}$ \\
\bottomrule
\end{tabular}
\end{table}

This section records implementation details.

\paragraph{Multi-scale encoder and pretraining.}
Each timestep contains 12 bilateral IMU channels: accelerometer $x,y,z$
followed by gyroscope $x,y,z$ for each foot. The three encoder streams operate
on 128, 256, and 512 samples at 100\,Hz, corresponding to 1.28, 2.56, and
5.12\,s. The streams use the same configuration but have independent
parameters. Unlike the pretraining model from which the encoder design was
adapted, the current multi-scale model does not use a sensor-location token or
location embedding.

Masked-sequence pretraining randomly removes 30\% of sensor readings and
reconstructs the original IMU values at the masked positions. A linear
projection maps each 12-dimensional bilateral timestep to the
128-dimensional model space. Six Transformer blocks, each with eight
self-attention heads and a $128\rightarrow512\rightarrow128$ feed-forward
network, process the sequence before a linear reconstruction head maps tokens
back to the 12-channel sensor space. The three temporal encoders use this same
configuration but are trained as separate parameter sets.

FoG detection uses focal
loss to address class imbalance. Batch size, learning rate, and other hyperparameters are tuned on the validation set. Training is performed on an NVIDIA A6000 GPU.

\begin{figure*}[ht!]
    \centering
    \includegraphics[width=\textwidth, trim=15cm 0cm 15cm 0cm, clip]{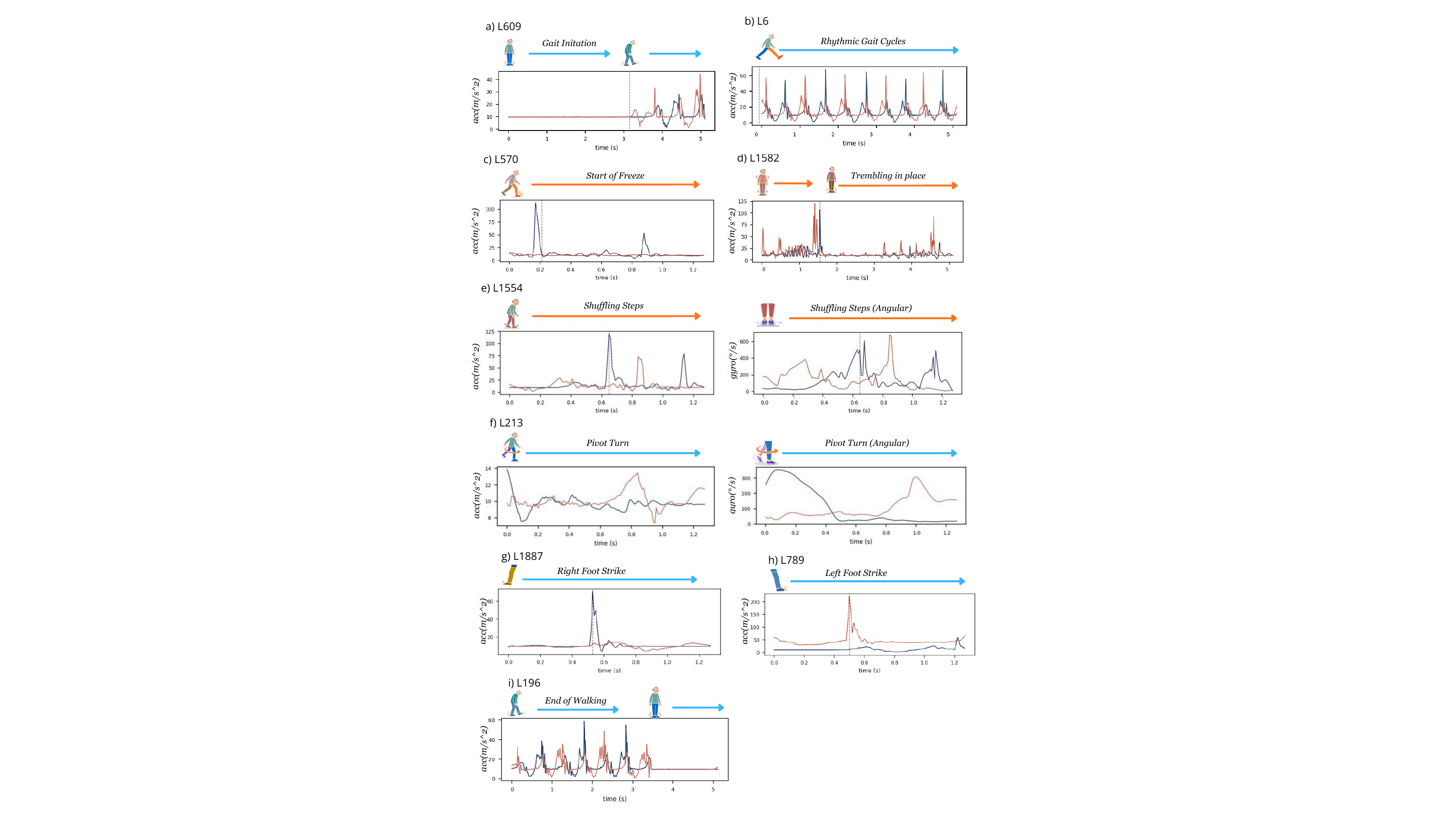}
    \caption{\textbf{Complete atlas of the nine discovered movement concepts.}
    Each panel shows a representative high-activation IMU window.
    The concepts span gait initiation, bilateral foot strikes, rhythmic gait
    cycles, start of freeze, trembling in place, shuffling steps, pivot turn,
    and end of walking. Acceleration is reported in $\mathrm{m/s^2}$ and
    angular velocity in ${}^{\circ}/\mathrm{s}$; red and blue denote the left
    and right foot, respectively.}
    \label{fig:app-atlas}
\end{figure*}

\paragraph{SAE training.} For SAE training, tokens from all three frozen encoder streams are pooled into
one training population. The SAE uses a 2048-feature dictionary with TopK
sparsity $k=32$. Decoder columns are normalized during training. After 20
epochs, 284 features satisfy the firing-rate criterion, and steering identifies 38 as model-relevant under $\Delta_f\geq0.10$. All 38 steering-selected features were inspected under the same characterization procedure; nine exhibited a consistent recurring movement pattern across de-duplicated high-activation windows and were assigned
movement-concept interpretations.

\section{Complete Movement-Concept Atlas}
\label{app:atlas}

Figure~\ref{fig:app-atlas} collects representative high-activation windows for
all nine final movement concepts, including the three concepts not analyzed in
detail in the main text: left-foot strike, right-foot strike, and end of
walking. Examples are selected using the window-level activation defined in
Eq.~\ref{eq:window-activation}. The complete set shows that the recovered
features span localized unilateral events, bilateral gait organization,
transitions between locomotor states, FoG-specific and PD-associated gait patterns, and coordinated rotational actions.

Table~\ref{tab:app-concepts} summarizes the final concept set together with
its dominant temporal scale and functional steering direction. The two
single-foot-strike features form a laterality-specific pair at the same short
temporal scale, while gait initiation and end of walking form complementary
long-context transitions. Positive steering of Start of Freeze, Trembling in Place, and Shuffling Steps raises \(P(\mathrm{FoG})\), indicating that the model treats both FoG-specific patterns and associated PD gait abnormalities as positive evidence for FoG.

\begin{table}[t]
\centering
\small
\caption{\textbf{Summary of the nine discovered movement concepts.}
$\uparrow$ indicates that positive steering increases $P(\mathrm{FoG})$;
$\downarrow$ indicates that it decreases $P(\mathrm{FoG})$.}
\label{tab:app-concepts}
\setlength{\tabcolsep}{5pt}
\begin{tabular}{llcc}
\toprule
Feature & Movement concept & Dominant scale & Steering \\
\midrule
L609  & Gait Initiation       & 5.12\,s & $\downarrow$ \\
L789  & Left Foot Strike      & 1.28\,s & $\downarrow$ \\
L1887 & Right Foot Strike     & 1.28\,s & $\downarrow$ \\
L6    & Rhythmic Gait Cycles  & 5.12\,s & $\downarrow$ \\
L570  & Start of Freeze       & 1.28\,s & $\uparrow$ \\
L1582 & Trembling in Place    & 5.12\,s & $\uparrow$ \\
L1554 & Shuffling Steps       & 1.28\,s & $\uparrow$ \\
L213  & Pivot Turn            & 1.28\,s & $\downarrow$ \\
L196  & End of Walking        & 5.12\,s & $\downarrow$ \\
\bottomrule
\end{tabular}
\end{table}

\section{Full Steering Effects}
\label{app:function}

\begin{figure*}[t!]
    \centering
    \includegraphics[width=\textwidth]{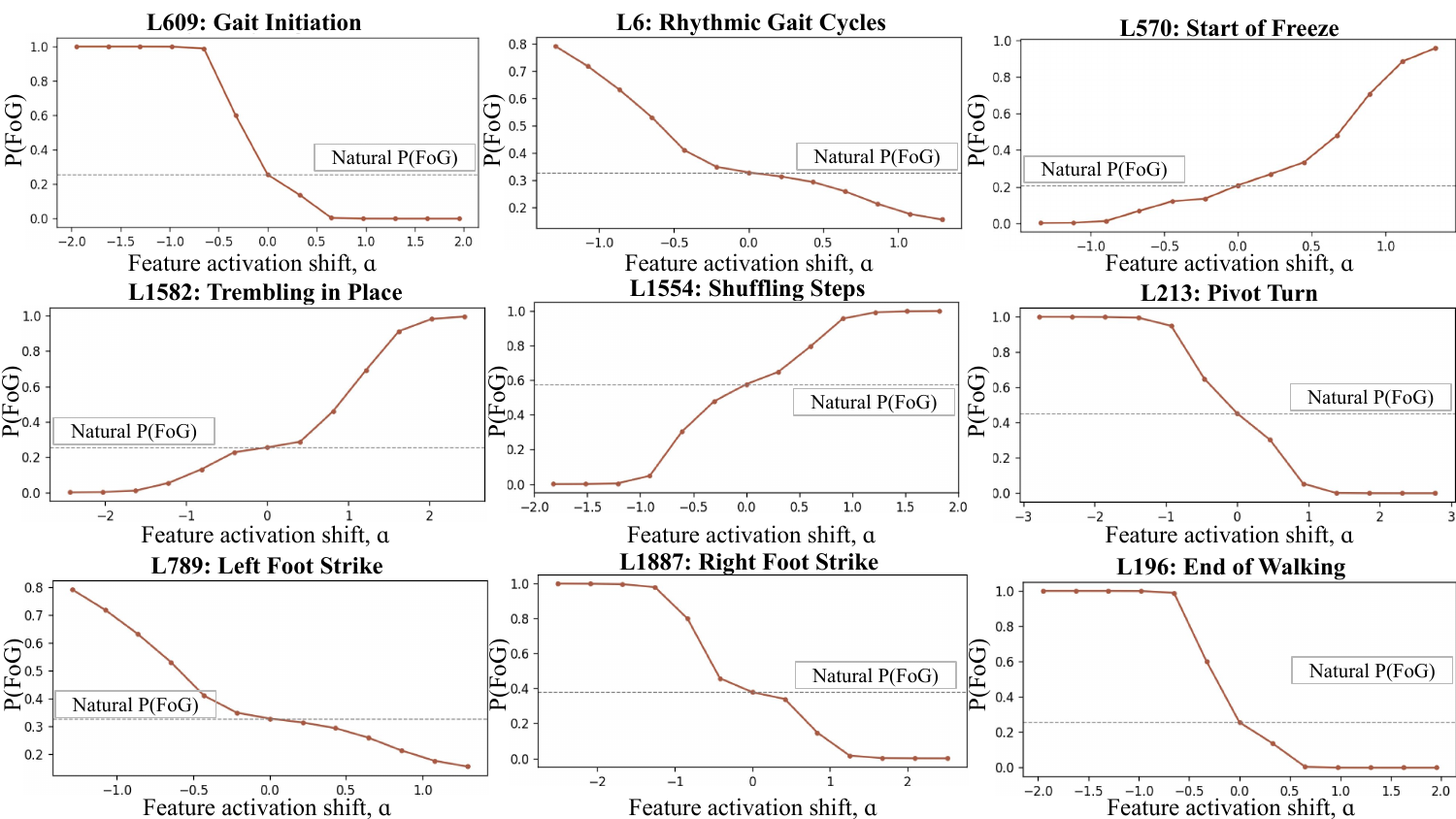}
    \caption{\textbf{Complete feature-steering responses for all nine movement
    concepts.}
    Each curve reports mean $P(\mathrm{FoG})$ over a fixed feature-active host
    set as the representation is shifted along the corresponding unit-norm SAE
    decoder direction. Dashed lines denote the natural, unmodified prediction.}
    \label{fig:app-steering}
\end{figure*}

\paragraph{Complete steering responses.}
Using the additive intervention defined in
Eq.~\ref{eq:feature-steering} and the steering settings in the main paper,
Figure~\ref{fig:app-steering} reports the complete responses for all nine
final concepts. The six concepts shown in the main paper are supplemented here
with L789, L1887, and L196. Positive steering of L570, L1582, and L1554 raises
the FoG prediction, whereas positive steering of L609, L789, L1887, L6, L213,
and L196 lowers it. Thus, Start of Freeze, Trembling in Place, and Shuffling Steps share a positive functional direction, although they span both FoG-specific phenomena and associated PD gait abnormalities, while the remaining movement concepts
provide negative evidence for FoG under the tested intervention.

The additional single-foot concepts provide a useful consistency check.
Strengthening either the left-foot-strike or right-foot-strike feature lowers
the FoG prediction, despite the features encoding opposite sides of the body.
End of Walking is also negative-direction, indicating that its functional role
is distinct from the FoG-transition feature L570 even though both involve a
change in locomotor state.

\paragraph{Temporal-scale preference.}
Scale identity is not provided to the shared SAE. We therefore quantify
temporal preference only after training. For feature $f$ and encoder scale
$s$, the scale-normalized firing rate is
\begin{equation}
r_{f,s}
=
\frac{1}{N_s}
\sum_{i:s_i=s}
\mathbf{1}[z_f(h_i)>0],
\label{eq:app-scale-rate}
\end{equation}
where $N_s$ is the number of evaluated tokens originating from scale $s$.
Normalizing by $N_s$ prevents the different sequence lengths from mechanically
determining the firing rate. The dominant scale is
\begin{equation}
s_f^\star
=
\operatorname*{arg\,max}_{s\in\{1.28,\,2.56,\,5.12\}}
r_{f,s}.
\label{eq:app-dominant-scale}
\end{equation}

Five of the nine concepts prefer the 1.28\,s stream and four prefer the
5.12\,s stream; none is dominated by the 2.56\,s stream
(Table~\ref{tab:app-concepts}). The short-context features include both
unilateral foot strikes, Start of Freeze, Shuffling Steps, and Pivot Turn,
whose defining structure is visible within a compact local interval. The
long-context features are Gait Initiation, Rhythmic Gait Cycles, Trembling in
Place, and End of Walking. For these concepts, interpretation depends on
extended before/after context, repeated gait cycles, or sustained movement
organization. The scale pattern therefore reflects the temporal extent needed
to distinguish different movement patterns rather than a simple
``transition versus state'' split.

\subsection{Random-Direction Intervention Ablation}
\label{app:random-steering}

The steering curves in Fig.~\ref{fig:app-steering} show that intervening
along the learned SAE directions can substantially change the FoG prediction.
To test whether these effects are specific to the learned feature directions, not a generic consequence of perturbing the hidden representation, we
compare each SAE intervention against norm-matched random directions.

\paragraph{Random-direction baseline.}
For each movement-concept feature $f$, we sample 1,000 random unit vectors
$r_j\in\mathbb{R}^{128}$, matching the normalization of the SAE decoder
direction $d_f$. We repeat the intervention using the same feature-active host
windows, the same naturally active token positions, and the same
feature-specific steering range as in Fig.~\ref{fig:app-steering}. Thus,
only the intervention direction is randomized:

\begin{equation}
\tilde{h}^{(j)}_{s,t}(\alpha)=
\begin{cases}
h_{s,t}+\alpha r_j, & z_f(h_{s,t})>0,\\
h_{s,t}, & z_f(h_{s,t})=0.
\end{cases}
\end{equation}

For each random direction, we compute the same steering effect

\begin{equation}
\Delta_f^{(j)}
=
\max_{\alpha\in\mathcal{A}_f}
\bar{p}^{(j)}_f(\alpha)
-
\min_{\alpha\in\mathcal{A}_f}
\bar{p}^{(j)}_f(\alpha),
\end{equation}

where $\bar{p}^{(j)}_f(\alpha)$ is the mean FoG probability over the fixed
host set after steering along $r_j$. We compare this null distribution with
the intervention range produced by the learned SAE direction shown in
Fig.~\ref{fig:app-steering}.

\begin{table}[t]
\centering
\caption{
\textbf{Random-direction intervention control.}
Random values summarize the distribution of steering effects from 1,000
norm-matched random directions. The SAE comparison indicates whether the
steering range shown in Fig.~\ref{fig:app-steering} exceeds the 95th
percentile or the maximum random effect.
}
\label{tab:random-steering}
\small
\begin{tabular}{llccc}
\toprule
Feature & Movement concept
& Random mean $\pm$ SD
& Random 95th
& SAE vs. random \\
\midrule
L609  & Gait Initiation
& $0.048\pm0.028$ & 0.085 & $>$ max \\

L789  & Left Foot Strike
& $0.066\pm0.056$ & 0.189 & $>$ max \\

L1887 & Right Foot Strike
& $0.109\pm0.134$ & 0.412 & $>$ max \\

L6    & Rhythmic Gait Cycles
& $0.196\pm0.175$ & 0.629 & $>$ 95th percentile \\

L570  & Start of Freeze
& $0.062\pm0.059$ & 0.200 & $>$ max \\

L1582 & Trembling in Place
& $0.133\pm0.165$ & 0.542 & $>$ max \\

L1554 & Shuffling Steps
& $0.166\pm0.188$ & 0.640 & $>$ max \\

L213  & Pivot Turn
& $0.062\pm0.056$ & 0.182 & $>$ max \\

L196  & End of Walking
& $0.022\pm0.039$ & 0.093 & $>$ max \\
\bottomrule
\end{tabular}
\end{table}

\paragraph{Results.}
The learned SAE directions produce substantially larger intervention effects
than random directions in the same representation space
(Table~\ref{tab:random-steering}). For eight of the nine movement concepts,
the steering range shown in Fig.~\ref{fig:app-steering} exceeds the largest
effect observed among 1,000 random directions. Rhythmic Gait Cycles (L6) is
the only exception: its learned-direction effect exceeds the 95th percentile
of the random distribution but not its maximum.

These results show that the large changes in FoG probability observed under
feature steering are not explained by arbitrary perturbations of comparable
magnitude. Rather, the effects are specific to the learned SAE directions,
providing additional evidence that the identified features are functionally
linked to the model's FoG decisions.
\section{Additional Movement Concepts Validation Analysis}
\label{app:crosssubject}

This section provides the complete construction of the cross-subject
validation tests summarized in Sec.~\ref{sec:pattern-discovery}. Movement
characterization uses 10 participants, while evaluation uses a disjoint set
of 13 participants. All probes and reference prototypes are constructed using
only the characterization subjects and are applied to the evaluation subjects
without refitting.

\paragraph{Raw-Signal Representation and Activation Labels.}
Each validation sample is one IMU window $w$. To evaluate the discovered
patterns independently of the SAE representation, we convert each window into
a 128-dimensional raw-signal descriptor. Let
$\va_L(t),\va_R(t)\in\R^3$ denote left- and right-foot acceleration and
$\bm{\omega}_L(t),\bm{\omega}_R(t)\in\R^3$ denote angular velocity. We compute
the magnitude of each signal, resample each trajectory to 32 points, and
concatenate them:
\begin{equation}
\begin{aligned}
\bm\phi(w)=\operatorname{concat}\Big(
&\mathcal{R}_{32}(\|a_L(t)\|_2),
 \mathcal{R}_{32}(\|\omega_L(t)\|_2),\\
&\mathcal{R}_{32}(\|a_R(t)\|_2),
 \mathcal{R}_{32}(\|\omega_R(t)\|_2)
\Big)
\in\mathbb{R}^{128},
\end{aligned}
\label{eq:app-raw-descriptor}
\end{equation}
where $\mathcal{R}_{32}$ denotes temporal resampling to 32 points.

For feature $f$, the window-level activation is defined in
Eq.~\ref{eq:window-activation}. We assign the binary activation label
\begin{equation}
y_f(w)=
\mathbf{1}[z_f(w)>0],
\label{eq:app-active-label}
\end{equation}
where $y_f=1$ denotes a feature-active window and $y_f=0$ an inactive window.
Inactive controls are sampled to match the temporal-scale distribution of the
active windows. Controls are not additionally matched by subject.

\paragraph{Recurrence.}
Recurrence measures whether the raw-signal structure associated with feature
activation remains detectable in subjects not used for characterization.
For each feature $f$, we fit a logistic-regression probe on characterization
windows using $\bm\phi(w)$ as input and $y_f(w)$ as the binary target:
\begin{equation}
q_f(w)
=
P\!\left(y_f=1\mid\bm\phi(w)\right).
\label{eq:app-recurrence-score}
\end{equation}
The fixed probe is then applied to evaluation-subject windows, producing one
continuous active-class score $q_f(w)$ for each window. Recurrence ROC-AUC is
computed from the held-out binary labels
$\{y_f(w_i)\}$ and corresponding probe scores
$\{q_f(w_i)\}$. A high recurrence AUC therefore indicates that active windows
in unseen subjects retain raw-signal structure that distinguishes them from
scale-matched inactive controls.

\paragraph{Prototype Consistency.}
Recurrence requires active windows to be separable from inactive controls but
does not require them to retain the characteristic signal form observed during
characterization. We therefore construct a fixed prototype for each feature
from its active characterization windows:
\begin{equation}
\vmu_f=
\frac{1}{|\mathcal{W}^{\mathrm{char}}_f|}
\sum_{w\in\mathcal{W}^{\mathrm{char}}_f}
\bm\phi(w),
\label{eq:app-prototype}
\end{equation}
where $\mathcal{W}^{\mathrm{char}}_f$ is the set of active characterization
windows for feature $f$.

No classifier is trained for this test. Each evaluation window receives one
scalar score given by its cosine similarity to the fixed prototype:
\begin{equation}
s_f(w)=
\frac{\bm\phi(w)^\top\vmu_f}
{\|\bm\phi(w)\|_2\|\vmu_f\|_2}.
\label{eq:app-prototype-score}
\end{equation}
Prototype-consistency ROC-AUC is computed from the same active/inactive labels
$\{y_f(w_i)\}$ and the similarity scores $\{s_f(w_i)\}$. A high AUC therefore
means that active windows from unseen subjects are consistently more similar
to the characterization-subject prototype than matched inactive controls.

\paragraph{Pairwise Specificity.}
Specificity tests whether different discovered features correspond to
distinguishable raw-signal patterns. For each pair of features
$(f_i,f_j)$, we collect their active characterization windows and assign the
binary feature-identity target
\begin{equation}
y_{ij}(w)=
\begin{cases}
1, & w\text{ is assigned to }f_i,\\
0, & w\text{ is assigned to }f_j.
\end{cases}
\label{eq:app-specificity-label}
\end{equation}
If a window activates both features, it is assigned to the feature with the
larger window-level activation $z_f(w)$. A logistic-regression probe takes
$\bm\phi(w)$ as input and predicts this feature identity. The fixed probe is then
applied to the disjoint evaluation subjects, producing one feature-identity
score per window. Pairwise ROC-AUC is computed from the held-out feature labels
and classifier scores. High specificity AUC therefore indicates that the two
feature-associated movement patterns remain distinguishable across subjects.

\paragraph{Per-Concept Cross-Subject Results.}
Table~\ref{tab:app-validation} reports the complete recurrence,
prototype-consistency, and coverage results. Recurrence AUC ranges from
0.90 to 0.99 (median 0.96), while prototype-consistency AUC ranges from
0.88 to 1.00 (median 0.95). The nine concepts occur in 15--23 participants
(median 19), indicating that the identified patterns are not confined to
isolated individuals.

\begin{table*}[t]
\centering
\small
\caption{\textbf{Cross-subject validation of the nine discovered movement
concepts.}
Recurrence AUC measures held-out separation between feature-active windows and
scale-matched inactive controls using a logistic probe. Prototype AUC measures
whether held-out active windows are more similar than controls to a fixed
prototype constructed from characterization subjects. Coverage denotes the
number of participants in whom the feature occurs.}
\label{tab:app-validation}
\setlength{\tabcolsep}{5pt}
\begin{tabular}{llccc}
\toprule
Feature & Movement concept & Recurrence AUC & Prototype AUC & Coverage \\
\midrule
L609  & Gait Initiation       & 0.96 & 0.93 & 22 \\
L789  & Left Foot Strike      & 0.99 & 1.00 & 18 \\
L1887 & Right Foot Strike     & 0.98 & 0.98 & 19 \\
L6    & Rhythmic Gait Cycles  & 0.97 & 0.97 & 23 \\
L570  & Start of Freeze       & 0.90 & 0.88 & 15 \\
L1582 & Trembling in Place    & 0.97 & 0.96 & 18 \\
L1554 & Shuffling Steps       & 0.94 & 0.91 & 16 \\
L213  & Pivot Turn            & 0.95 & 0.94 & 20 \\
L196  & End of Walking        & 0.96 & 0.95 & 21 \\
\bottomrule
\end{tabular}
\end{table*}

The two unilateral foot-strike concepts achieve particularly high recurrence
and prototype consistency, indicating that their feature-associated signal
structure transfers strongly across participants. Start of Freeze has the
lowest recurrence and prototype AUCs among the nine concepts
(0.90 and 0.88, respectively), but remains well above chance in both tests.
This variation suggests that some movement concepts have more stereotyped
raw-signal realizations than others.

\paragraph{Pairwise Concept Relationships.}
Table~\ref{tab:app-specificity} reports held-out specificity AUC for all
36 concept pairs. The overall median is 0.875. Same-scale pairs have a median
AUC of 0.915, compared with 0.855 for cross-scale pairs. Because concepts from
the same encoder stream share the same temporal context, their stronger
separation indicates that specificity is not explained simply by scale
identity.

\begin{table*}[t]
\centering
\scriptsize
\caption{\textbf{Held-out pairwise specificity AUC among the nine movement
concepts.}
Each entry measures how well a logistic probe distinguishes active windows of
the corresponding feature pair in the disjoint evaluation subjects. Windows
activating both features are assigned to the feature with the stronger
window-level activation.}
\label{tab:app-specificity}
\setlength{\tabcolsep}{3.2pt}
\begin{tabular}{lrrrrrrrrr}
\toprule
 & L609 & L789 & L1887 & L6 & L570 & L1582 & L1554 & L213 & L196 \\
\midrule
L609  & --   & 0.82 & 0.80 & 0.78 & 0.88 & 0.99 & 0.84 & 0.86 & 0.74 \\
L789  & 0.82 & --   & 1.00 & 0.84 & 0.90 & 0.94 & 0.92 & 0.97 & 0.80 \\
L1887 & 0.80 & 1.00 & --   & 0.85 & 0.91 & 0.95 & 0.93 & 1.00 & 0.81 \\
L6    & 0.78 & 0.84 & 0.85 & --   & 0.93 & 0.89 & 0.82 & 0.87 & 0.76 \\
L570  & 0.88 & 0.90 & 0.91 & 0.93 & --   & 0.86 & 0.72 & 0.98 & 0.85 \\
L1582 & 0.99 & 0.94 & 0.95 & 0.89 & 0.86 & --   & 0.70 & 0.90 & 1.00 \\
L1554 & 0.84 & 0.92 & 0.93 & 0.82 & 0.72 & 0.70 & --   & 0.88 & 0.86 \\
L213  & 0.86 & 0.97 & 1.00 & 0.87 & 0.98 & 0.90 & 0.88 & --   & 0.89 \\
L196  & 0.74 & 0.80 & 0.81 & 0.76 & 0.85 & 1.00 & 0.86 & 0.89 & -- \\
\bottomrule
\end{tabular}
\end{table*}

The pairwise structure provides additional evidence that the recovered
features encode specific movement organization rather than only coarse
temporal context. Left and Right Foot Strike are perfectly separable
(AUC $=1.00$) despite representing the same type of localized event at the
same 1.28\,s scale. Their separation therefore reflects laterality in the
underlying movement rather than temporal-scale differences. Other highly
separable same-scale pairs include Right Foot Strike versus Pivot Turn
(AUC $=1.00$) and Start of Freeze versus Pivot Turn (AUC $=0.98$).

Conversely, several of the least-separable pairs correspond to movements with
related signal structure. Trembling in Place versus Shuffling Steps has AUC
$0.70$, and Start of Freeze versus Shuffling Steps has AUC $0.72$; all three capture disrupted gait patterns relevant to FoG detection. Gait Initiation versus
End of Walking has AUC $0.74$, while Rhythmic Gait Cycles versus End of
Walking has AUC $0.76$, consistent with their shared locomotor structure.
The specificity matrix therefore reflects graded relationships among movement
concepts rather than arbitrary separability of feature identities.

\section{SAE Information Loss and Configuration Sensitivity}
\label{app:sae}

\paragraph{Representation fidelity.}
Before interpreting the learned sparse features, we verify that the SAE
preserves the encoder representation and the downstream FoG prediction.
For a validation token $h$ and its SAE reconstruction $\hat{h}$, we measure
variance explained as
\begin{equation}
\mathrm{VE}
=
1-
\frac{
  \mathbb{E}\!\left[\|\vh-\hat{\vh}\|_2^2\right]
}{
  \mathbb{E}\!\left[\|\vh-\vmu_h\|_2^2\right]
},
\end{equation}
where $\mu_h$ is the mean validation token representation.
The SAE used in the main analysis, with
$d_{\mathrm{SAE}}=2048$ and TopK sparsity $k=32$, the SAE achieves a
validation reconstruction $R^2$ of 0.9976.

We additionally measure whether this reconstruction preserves the downstream
computation performed by the frozen FoG model. We replace the original encoder
tokens with their SAE reconstructions and propagate them through the unchanged
pooling, fusion, and classification layers. The original and reconstructed
FoG probabilities have Pearson correlation $r=0.9999$, with a mean absolute
probability difference of 0.0013. FoG-detection F1 changes by approximately
$-0.0001$. Thus, although the sparse reconstruction is not exact in
representation space, it preserves nearly all task-relevant information at
the analyzed layer.

\paragraph{Sensitivity to SAE configuration.}
We examine the sensitivity of the decomposition to dictionary width
$d_{\mathrm{SAE}}$ and TopK sparsity $k$. We train SAEs with
$d_{\mathrm{SAE}}\in\{512,1024,2048\}$ and $k\in\{16,32\}$ where available,
using the same training procedure. Table~\ref{tab:sae-sensitivity} reports
validation reconstruction $R^2$ and the number of alive features, where a
feature is considered alive if its firing rate exceeds $10^{-3}$.

\begin{table}[t]
    \centering
    \caption{
    \textbf{Sensitivity to SAE configuration.}
    $R^2$ measures validation reconstruction fidelity.
    Alive features satisfy the firing-rate criterion $r_f>10^{-3}$.
    }
    \label{tab:sae-sensitivity}
    \begin{tabular}{cccc}
        \toprule
        $d_{\mathrm{SAE}}$ & $k$ & Validation $R^2$ & Alive features \\
        \midrule
        512  & 32 & 0.9973 & 245 / 512 \\
        1024 & 16 & 0.9931 & 198 / 1024 \\
        1024 & 32 & 0.9974 & 258 / 1024 \\
        2048 & 16 & 0.9923 & 194 / 2048 \\
        \textbf{2048} & \textbf{32} & \textbf{0.9976} & \textbf{284 / 2048} \\
        \bottomrule
    \end{tabular}
\end{table}

At matched dictionary widths, $k=32$ consistently provides both higher
reconstruction fidelity and a larger set of active features than $k=16$.
For $d_{\mathrm{SAE}}=1024$, increasing $k$ from 16 to 32 raises
$R^2$ from 0.9931 to 0.9974 and the number of alive features from 198 to
258. The same pattern holds at $d_{\mathrm{SAE}}=2048$, where $R^2$
increases from 0.9923 to 0.9976 and the number of alive features from 194
to 284. We therefore use $k=32$ for the main analysis.

Among the $k=32$ configurations, reconstruction fidelity is already near
saturation: validation $R^2$ changes only from 0.9973 at
$d_{\mathrm{SAE}}=512$ to 0.9974 at 1024 and 0.9976 at 2048.
Dictionary width therefore has little effect on information preservation in
this range. Instead, increasing the width provides additional capacity for
separating recurring representation patterns. During inspection of the
512-dimensional SAE, we observed greater qualitative redundancy, with
multiple features responding to similar movement patterns, suggesting that
the narrower dictionary can allocate multiple features to closely related
structure.

We use $d_{\mathrm{SAE}}=2048$, $k=32$ as a high-fidelity, overcomplete
operating point for the movement-concept analysis. This setting retains the
near-lossless reconstruction observed at smaller widths while providing
greater dictionary capacity and the largest number of alive features among
the tested configurations. Importantly, our choice is not based on a claim
that $d_{\mathrm{SAE}}=2048$ is uniquely optimal: the 1024- and
2048-dimensional SAEs have nearly identical reconstruction fidelity.
Rather, the wider dictionary is used to reduce capacity constraints on the
feature decomposition while preserving downstream model behavior.

The qualitative redundancy observation was not converted into a systematic
semantic-overlap metric across all SAE configurations. We therefore interpret
this experiment as a configuration-sensitivity analysis rather than evidence
for a unique optimal dictionary size. The main analysis only requires that
the selected SAE provide a sparse, high-fidelity decomposition from which
functionally relevant and reproducible movement concepts can be identified.

\section{Reproducibility Across Datasets and Model Configurations}
\label{app:robustness}

This section evaluates whether FoGConcept identifies similar human-understandable movement concepts on other datasets and under different model configurations. We repeat the complete pipeline with changes on: new datasets, wider timescales, and a deeper Transformer backbone.

\subsection{Experimental Settings}
\label{app:robustness-settings}

Table~\ref{tab:robustness-configs} summarizes the evaluated configurations. The primary VCU model uses bilateral ankle IMUs sampled at 100Hz and three temporal streams of 1.28, 2.56, and 5.12s. FoG-STAR~\citep{fogstar} provides an independent cohort of 22 PD participants recorded using different IMU hardware and a source sampling rate of 60Hz.

The wider-context models retain three streams but double all input durations from 1.28/2.56/5.12s to 2.56/5.12/10.24s. The deeper models increase each temporal encoder from 6 to 12 Transformer layers while keeping the hidden dimension (128), eight attention heads, feed-forward width (512), dropout, classifier, temporal inputs, training protocol, LOSO evaluation, SAE configuration ($d_{\mathrm{SAE}}=2048$, $k=32$), steering settings, and characterization procedure fixed.

\begin{table*}[t]
\centering
\small
\caption{Configurations used for reproducibility analysis.}
\label{tab:robustness-configs}
\setlength{\tabcolsep}{5pt}
\begin{tabular}{lccccc}
\toprule
Configuration & Subjects & Placement & Source rate & Layers & Temporal contexts \\
\midrule
VCU, primary & 23 & Bilateral ankle & 100Hz & 6 & 1.28 / 2.56 / 5.12s \\
FoG-STAR, ankle & 22 & Bilateral ankle & 60Hz & 6 & 1.28 / 2.56 / 5.12s \\
FoG-STAR, back & 22 & Lower back & 60Hz & 6 & 1.28 / 2.56 / 5.12s \\
VCU, wider & 23 & Bilateral ankle & 100Hz & 6 & 2.56 / 5.12 / 10.24s \\
FoG-STAR, wider & 22 & Bilateral ankle & 60Hz & 6 & 2.56 / 5.12 / 10.24s \\
VCU, 12-layer & 23 & Bilateral ankle & 100Hz & 12 & 1.28 / 2.56 / 5.12s \\
FoG-STAR, 12-layer & 22 & Bilateral ankle & 60Hz & 12 & 1.28 / 2.56 / 5.12s \\
\bottomrule
\end{tabular}
\end{table*}

\subsection{Independent-Dataset Reproduction on FoG-STAR}
\label{app:fogstar}

We first repeat the complete FoGConcept procedure on the bilateral ankle signals from FoG-STAR. The FoG-STAR detector and SAE are trained independently, and we do the same steering-based feature selection and movement characterization.

All nine movement-concepts from the primary VCU analysis recur on FoG-STAR. All nine retain the same direction of their functional effect on $P(\mathrm{FoG})$ as in the primary VCU analysis.

\begin{figure*}[t]
    \centering
    \includegraphics[width=\textwidth, trim=15cm 0cm 15cm 0cm, clip]{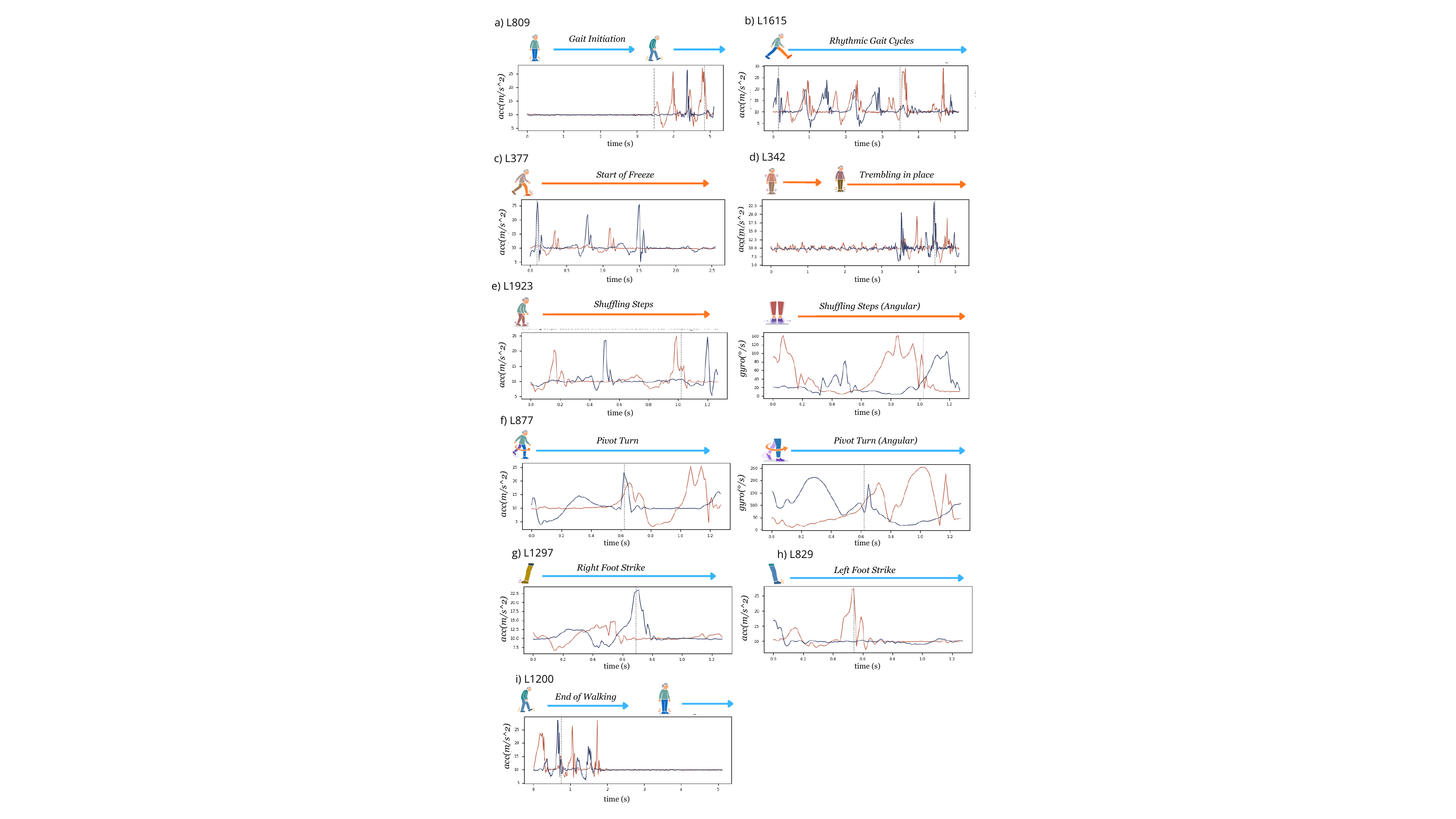}
    \caption{\textbf{Movement concepts independently recovered on FoG-STAR.}
    Representative high-activation windows for all nine movement-concepts recovered from the independently trained FoG-STAR model. Red and navy denote the left and right ankle, respectively. Vertical dashed lines indicate the token of maximum feature activation where shown. Despite changes in participants, recording sites, hardware, source sampling rate, and clinical protocol, the same nine movement-concepts identified in the primary VCU analysis recur on FoG-STAR.}
    \label{fig:fogstar-atlas}
\end{figure*}

Figure~\ref{fig:fogstar-atlas} shows representative high-activation windows for the nine independently recovered FoG-STAR concepts.

Table~\ref{tab:concept-recurrence} summarizes concept recurrence across all bilateral-ankle configurations.

\begin{table*}[t]
\centering
\small
\caption{\textbf{Movement-concept recurrence across datasets and model configurations.}
$\checkmark$ denotes recovery of the same movement-concept family, and -- denotes that the concept was not recovered as a distinct interpretable concept. Wider models use 2.56/5.12/10.24s contexts.}
\label{tab:concept-recurrence}
\resizebox{\textwidth}{!}{
\begin{tabular}{lcccccc}
\toprule
Movement concept
& VCU primary
& FoG-STAR
& VCU wider
& FoG-STAR wider
& VCU 12-layer
& FoG-STAR 12-layer \\
\midrule
Gait Initiation          & $\checkmark$ & $\checkmark$ & $\checkmark$ & $\checkmark$ & $\checkmark$ & $\checkmark$ \\
Left Foot Strike         & $\checkmark$ & $\checkmark$ & $\checkmark$ & $\checkmark$ & --           & -- \\
Right Foot Strike        & $\checkmark$ & $\checkmark$ & $\checkmark$ & $\checkmark$ & --           & -- \\
Rhythmic Gait Cycles     & $\checkmark$ & $\checkmark$ & $\checkmark$ & $\checkmark$ & $\checkmark$ & $\checkmark$ \\
Start of Freeze          & $\checkmark$ & $\checkmark$       & $\checkmark$ & $\checkmark$ & $\checkmark$ & $\checkmark$ \\
Trembling in Place       & $\checkmark$ & $\checkmark$ & $\checkmark$ & $\checkmark$ & $\checkmark$ & $\checkmark$ \\
Shuffling Steps          & $\checkmark$ & $\checkmark$       & $\checkmark$ & $\checkmark$ & $\checkmark$ & $\checkmark$ \\
Pivot Turn               & $\checkmark$ & $\checkmark$ & $\checkmark$ & $\checkmark$ & $\checkmark$ & $\checkmark$ \\
End of Walking           & $\checkmark$ & $\checkmark$ & $\checkmark$ & $\checkmark$ & $\checkmark$ & $\checkmark$ \\
\midrule
Original families recovered
                        & 9/9 & 9/9 & 9/9 & 9/9 & 7/9 & 7/9 \\
Additional concept      & -- & -- & -- & -- & Unilateral Freezing & Standing Still\\
\bottomrule
\end{tabular}}
\end{table*}

\subsection{Generalization to Wider Temporal Contexts}
\label{app:wider}

We next test whether the concept set depends on the exact temporal scale. We analyze for 10.24s input windows. Both wider-context models on FoG-STAR and the VCU dataset recover all nine movement-concepts from the primary analysis. This result indicates that the recovered movement vocabulary is not tied to the exact 1.28/2.56/5.12s timescales. At the same time, individual concepts need not occupy a unique temporal stream, as common movement patterns can remain recognizable across multiple contexts, whereas other concepts become most interpretable only when sufficient local or sustained temporal information is available.

The 10.24s stream is used here as a generalization test. FoG is clinically episodic and most commonly lasts only a few seconds, although longer episodes can occur~\citep{nutt2011freezing}. Long analysis windows have been explored in prior FoG-detection work, including windows up to 10s~\citep{moore2013autonomous}. However, longer windows can smooth short FoG events and are less well suited to responsive online intervention; prior analyses found that windows above approximately 4\,s can reduce sensitivity to short events and increase detection delay, with shorter windows preferred when rapid detection is prioritized~\citep{kwon2014practical}. Real-time FoG systems accordingly often operate on shorter contexts windows. We therefore retain the shorter temporal hierarchy in the primary model and use 10.24s only to test whether the interpretation procedure remains stable under substantially wider context.

\subsection{Generalization to Backbone Depth}
\label{app:depth}

We next increase the temporal encoder depth from 6 to 12 Transformer layers while holding the remaining model, training, SAE, steering, and evaluation settings fixed. Both 12-layer models on VCU and the FoG-STAR dataset recover seven of the nine movement-concepts from the primary analysis. Gait Initiation, Rhythmic Gait Cycles, Start of Freeze, Trembling in Place, Shuffling Steps, Pivot Turn, and End of Walking recur on both datasets, while the two unilateral Foot Strike concepts are not recovered as distinct concept families.

The deeper models also expose additional interpretable structure. On VCU, one feature consistently captures \emph{Unilateral Freezing}: one foot exhibits repeated oscillatory or stepping attempts while the contralateral foot remains planted or nearly motionless. This strongly asymmetric organization is consistent with the clinical observation that FoG can preferentially affect one foot~\citep{nutt2011freezing}; recent slow-motion videography similarly describes a ``Fulcrum Sign,'' in which one foot remains relatively stationary while the other exhibits repeated stepping and trembling during FoG~\citep{wali2026slow}.

On FoG-STAR, an additional feature captures \emph{Standing Still}. Across its high-activation windows, both feet remain nearly stationary without freezing episodes, indicating voluntary stops during walking. Maximal feature activations occur within the stationary interval and not at the subsequent onset of movement, so we therefore distinguish This is also different from Gait Initiation, whose defining structure is the transition from standing into repeated stepping. Figure~\ref{fig:additional-concepts} shows representative examples of these two additional concepts.

The deeper-backbone results therefore show both stability and representation-dependent variation. Most primary movement-concept families remain interpretable after doubling backbone depth, while the altered representation can expose additional clinically meaningful distinctions. We present FoGConcept as a reproducible concept-discovery procedure, and we do not expect it to return a complete and unique concept vocabulary for every model.

\subsection{Sensor-Placement Boundary}
\label{app:placement}

FoG-STAR additionally provides a simultaneously recorded lower-back IMU, allowing us to examine a substantially different sensing representation while holding participants and annotations fixed. We repeat the same detector, SAE, and steering procedure using the lower-back signal alone.

We treat this experiment as a sensing-boundary analysis rather than requiring concept identity with the bilateral-ankle model. Several concepts in the primary analysis, such as Left and Right Foot Strike and bilateral gait coordination, depend directly on information from the two feet and are therefore not expected to be represented identically by a single trunk sensor. The lower-back experiment nevertheless produces a valid sparse decomposition and functionally relevant steering features (Table~\ref{tab:sae-robustness}), while detection performance is lower than with bilateral ankle sensing (Table~\ref{tab:detection-robustness}). This result further motivates distinguishing reproducibility of the FoGConcept procedure from invariance of the exact movement-concept vocabulary across sensor placements.

\subsection{SAE Fidelity and Steering Diagnostics}
\label{app:robustness-sae}

Table~\ref{tab:sae-robustness} summarizes SAE reconstruction and steering statistics across all evaluated configurations. Across the six new robustness settings, token reconstruction $R^2$ ranges from 0.988 to 0.9998, reconstructed and original FoG probabilities have Pearson correlation $r=0.9989$--$1.000$, and replacing encoder tokens with their SAE reconstructions changes weighted F1 only negligibly. Thus, differences in recovered concepts are not attributable to poor SAE reconstruction.

\begin{table*}[t]
\centering
\scriptsize
\caption{\textbf{SAE fidelity and steering statistics across configurations.} ``Selected'' denotes features with $\Delta_f\geq0.10$.}
\label{tab:sae-robustness}
\resizebox{\textwidth}{!}{
\begin{tabular}{lccccccc}
\toprule
 & FoG-STAR ankle
 & FoG-STAR back
 & FoG-STAR wider
 & FoG-STAR 12L
 & VCU primary
 & VCU wider
 & VCU 12L \\
\midrule
Alive features
& 388 & 267 & 317 & 470 & 284 & 625 & 420 \\

Selected ($\Delta_f\geq0.10$)
& 96 & 47 & 100 & 27 & 38 & 73 & 28 \\

Above random 95th percentile
& 45 & 29 & 48 & 27 & 9/9 concepts & 38 & 27 \\

Above random maximum
& 10 & 8 & 12 & 19 & 8/9 concepts & 13 & 22 \\

Dominant scale (short / middle / long)
& 27 / 32 / 37
& 14 / 8 / 25
& 36 / 24 / 40
& 14 / 2 / 11
& 5 / 0 / 4
& 28 / 16 / 29
& 11 / 0 / 17 \\

Token $R^2$
& 0.988
& 0.991
& 0.9885
& 0.9996
& 0.9976
& 0.9889
& 0.9998 \\

$P(\mathrm{FoG})$ correlation
& 0.9991
& 0.9998
& 0.9991
& 1.000
& 0.9999
& 0.9989
& 1.000 \\

F1, original $\rightarrow$ reconstructed
& 0.936 $\rightarrow$ 0.935
& 0.887 $\rightarrow$ 0.886
& 0.923 $\rightarrow$ 0.925
& 0.890 $\rightarrow$ 0.890
& $\Delta\approx-0.0001$
& 0.956 $\rightarrow$ 0.955
& 0.944 $\rightarrow$ 0.944 \\
\bottomrule
\end{tabular}}
\end{table*}

\subsection{Detection Performance Under Different Settings}
\label{app:robustness-detection}

Table~\ref{tab:detection-robustness} reports LOSO detection performance for the additional configurations. Epoch and decision threshold are selected using validation subjects only. These experiments are intended primarily to evaluate concept reproducibility, not to optimize detection accuracy.

\begin{table*}[t]
\centering
\small
\caption{\textbf{FoG detection under additional dataset and model configurations.} F1 is reported across LOSO test subjects as mean$\pm$SD where available.}
\label{tab:detection-robustness}
\setlength{\tabcolsep}{6pt}
\begin{tabular}{lccc}
\toprule
Configuration
& F1
& AUROC\\
\midrule
FoG-STAR ankle
& $0.831\pm0.142$
& 0.871\\

FoG-STAR back
& $0.755\pm0.178$
& 0.830\\

FoG-STAR wider
& $0.855\pm0.120$
& 0.873\\

FoG-STAR 12-layer
& $0.808\pm0.137$
& 0.832\\

VCU wider
& $0.881\pm0.110$
& 0.931\\

VCU 12-layer
& $0.853\pm0.113$
& 0.865\\
\bottomrule
\end{tabular}
\end{table*}

\appendix

\end{document}